%% file: main.tex
\pdfoutput=1

\documentclass[11pt]{article}

\usepackage{EMNLP2024}

\usepackage{times}
\usepackage{latexsym}

\usepackage[T1]{fontenc}

\usepackage[utf8]{inputenc}

\usepackage{microtype}
\usepackage{authblk}
\usepackage{inconsolata}
\usepackage{graphicx}
\usepackage{booktabs}
\usepackage{amsmath, amsthm,bm, amssymb}
\usepackage{natbib}
\usepackage{multirow}
\usepackage{url}
\usepackage{subfigure}
\usepackage{subcaption}
\newtheorem{definition}{Definition}
\title{Explaining Graph Neural Networks with Large Language Models:\\A Counterfactual Perspective for Molecular Property Prediction}

\author{
    \textbf{Yinhan He}\textsuperscript{†}, 
    \textbf{Zaiyi Zheng}\textsuperscript{†} 
    \textbf{Patrick Soga}\textsuperscript{†} \\ 
    \textbf{Yaochen Zhu}\textsuperscript{†} 
    \textbf{Yushun Dong}\textsuperscript{§} 
    \textbf{Jundong Li}\textsuperscript{†}
}
\affil{\textsuperscript{†}University of Virginia, Charlottesville, VA, USA \vspace{-4mm}}
\affil{\textsuperscript{§}Florida State University, Tallahassee, FL, USA \vspace{-4mm}}
\affil{\texttt{\{nee7ne, sjc4fq, zqe3cg, uqp4qh, jl6qk\}@virginia.edu}, \texttt{yd24f@fsu.edu}}

\begin{document}
\maketitle
\begin{abstract}
    \input{Sections/0-abstract}

\end{abstract}

\section{Introduction}\label{sec:introduction}
\input{Sections/1-introduction}

\section{Preliminaries}\label{sec:preliminaries}
\input{Sections/2-preliminaries}

\section{Data Construction}\label{sec:data_prepare}
\input{Sections/3-data_preparation}

\section{Methodology}\label{sec:methods}
\input{Sections/4-method}

\section{Experiments}\label{sec:experiments}
\input{Sections/5-experiments}

\section{Conclusion}\label{sec:conclusion}
\input{Sections/7-conclusions}
\section{Acknowledgement}
\input{Sections/9-acknowledgement}
\clearpage
\section{Limitations}\label{sec:limitations}
\input{Sections/8-limitations}

\section{Ethics Statement}
\input{Sections/Ethics}

\vspace{-15pt}
\bibliography{main}
\clearpage
\label{sec:appendix}
\input{Appendix/supplementary_material}

\end{document}

%% file: Sections/0-abstract.tex
In recent years, Graph Neural Networks (GNNs) have become successful in molecular property prediction tasks such as toxicity analysis. However, due to the black-box nature of GNNs, their outputs can be concerning in high-stakes decision-making scenarios, e.g., drug discovery.
%
Facing such an issue, Graph Counterfactual Explanation (GCE) has emerged as a promising approach to improve GNN transparency. However, current GCE methods usually fail to take domain-specific knowledge into consideration, which can result in outputs that are not easily comprehensible by humans.
To address this challenge, we propose a novel GCE method, LLM-GCE, to unleash the power of large language models (LLMs) in explaining GNNs for molecular property prediction.
%
Specifically, we utilize an autoencoder to generate the counterfactual graph topology from a set of counterfactual text pairs (CTPs) based on an input graph. Meanwhile, we also incorporate a CTP dynamic feedback module to mitigate LLM hallucination, which provides intermediate feedback derived from the generated counterfactuals as an attempt to give more faithful guidance.
Extensive experiments demonstrate the superior performance of LLM-GCE.
 Our code is released on ~\href{https://github.com/YinhanHe123/new\_LLM4GNNExplanation}{https://github.com/YinhanHe123/new\_LLM4G\break NNExplanation}.

%% file: Sections/1-introduction.tex
Molecular property prediction has attracted increasing attention in recent years, where Graph Neural Networks (GNNs) have achieved significant success in the related downstream tasks, such as drug discovery~\cite{xiong2021graph} and toxicity analysis~\cite{cremer2023equivariant}.
%
%
However, GNNs are typically considered as black-box models, making it difficult for users to understand how a given prediction is derived.
Such a lack of explainability 
brings obstacles against their broader real-world applications to understand molecular properties.

\begin{figure}[t]
\centering
\includegraphics[width=0.5\textwidth]{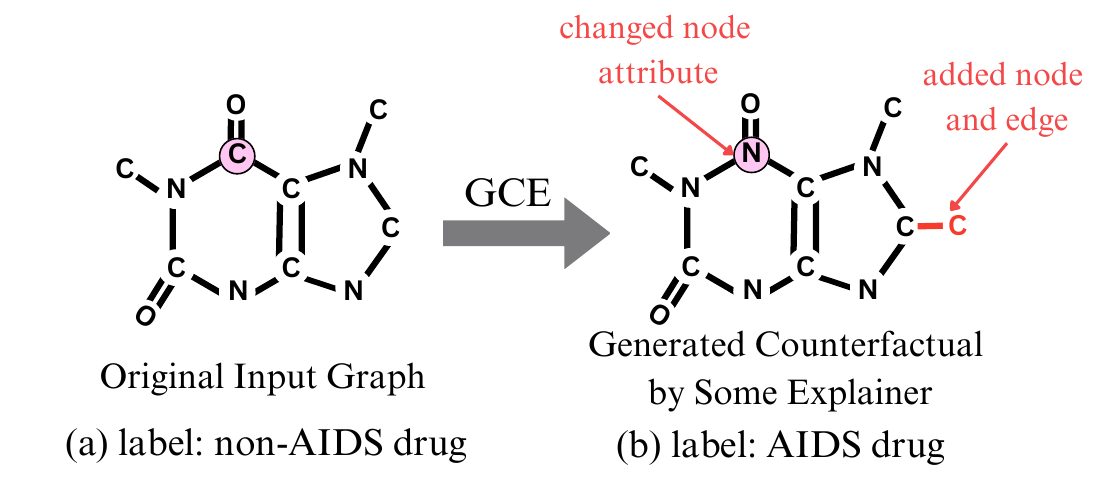}
\vspace{-0.3in}
\caption{
Example of graph counterfactual explanation. 
(a) the original input molecule graph; (b) an example counterfactual graph. 
}
\vspace{-0.15in}
\label{fig:counterfactual_illustration}
\end{figure}
Facing such an issue, a series of approaches have been proposed to explain the predictions of GNNs, where graph counterfactual explanations (GCE) have become a prevalent approach in recent years~\cite{ying2019gnnexplainer, lucic2022cf, ma2022clear, zhang2023regexplainer}. Specifically, GCE aims to identify the minimum modification over a given graph, such that a trained GNN yields a desired prediction for the post-modified graph. Here, the graph with the identified modification is called the \textit{counterfactual graph}, or simply counterfactual for short. The identified modifications may involve adding or removing nodes and/or edges, as well as altering the node/edge attributes. For instance, given an undesired (non-AIDS drug) molecule as in Fig.~\ref{fig:counterfactual_illustration}(a), a GCE method may generate modifications to produce a molecule as a ``desired'' graph (i.e., predicted as an AIDS-drug by the GNN model) as shown in Fig.~\ref{fig:counterfactual_illustration}(b).

However, existing GCE models have two significant limitations: \emph{(i) Incomprehensible Counterfactual Optimization}.
Most GCE models are optimized to generate counterfactuals either through heuristic methods, such as random walk~\cite{huang2023global}, or black-box deep learning methods~\cite{ying2019gnnexplainer, bajaj2021robust, lucic2022cf, ma2022clear, tan2022learning}, which fail to involve any human-interpretable knowledge in optimizing the counterfactuals. \emph{(ii) Lack of Domain Knowledge}. Most current GCE methods do not consider any domain-specific knowledge~\cite{ying2019gnnexplainer, bajaj2021robust, lucic2022cf, ma2022clear, tan2022learning} 
thus the generated counterfactuals may not be realistic in real-world contexts. Continuing the example shown in Fig.~\ref{fig:counterfactual_illustration}, although the generated counterfactual Fig.~\ref{fig:counterfactual_illustration}(b) is classified as in the desired class, it is not chemically stable since it violates
the valence bond theory~\citep{lewis1933chemical}.
To handle the above limitations, large language models (LLMs)~\citep{radford2018improving,wu2024usable} are ideal for addressing these limitations due to their ability to \emph{(i)} generate comprehensible natural language texts, \emph{(ii)} make the counterfactual optimization process human-interpretable, and \emph{(iii)} leverage inherent domain knowledge from extensive pretraining to produce realistic counterfactuals.
However, harnessing LLMs to improve counterfactual explanation generation faces challenges: \emph{(i)} 
there exists a natural mismatch between texts (sequential data) and graph structures~\cite{wang2024can, li2023survey};
and \emph{(ii)} LLMs may hallucinate, i.e., generate seemingly plausible while incorrect information~\cite{huang2023survey}.

\vspace{-2pt}
To handle these challenges, we propose a novel framework: LLM-GCE (\underline{\textbf{L}}arge \underline{\textbf{L}}anguage \underline{\textbf{M}}odels guided \underline{\textbf{G}}raph \underline{\textbf{C}}ounterfactual \underline{\textbf{E}}xplainer). Specifically, to mitigate the first challenge, instead of directly generating the counterfactual graphs by LLMs, we utilize 
a counterfactual autoencoder (CA) to construct counterfactual graph structures based on the the  text pairs (TPs) and counterfactual text pairs (CTPs) given by LLMs. 
To tackle the second challenge, the hallucination, we design a CTP dynamic feedback module enlightened by ~\citet{madaan2024self}
to update CTPs iteratively based on previously generated counterfactuals.


\vspace{-1pt}
Our main contributions are summarized as follows: \textbf{\textit{(i)}} 
 \underline{\textit{\textbf{Dataset Construction.}}} We collect LLM-generated text pairs over five molecule datasets. They not only support our empirical evaluations but also facilitate future studies for researchers in this field. \textbf{\textit{(ii)}} \underline{\textit{\textbf{Algorithmic Design.}}}  We propose a novel LLM-GCE framework that learns to generate graph counterfactual explanations under the guidance of an LLM. LLM-GCE unlocks LLM's strong reasoning ability in GCE by addressing hallucinations and graph structure inference limitations. \textbf{\textit{(iii)}} \underline{\textit{\textbf{Experimental Evaluation.}}} We conduct extensive experiments on multiple real-world datasets, validating the effectiveness of LLM in generating more feasible counterfactuals while providing a comprehensive optimization trajectory.

%% file: Sections/2-preliminaries.tex
In this section, we introduce the problem settings for GCE and the evaluation metrics used. We denote a molecule graph of $m$ nodes (atoms) as $G=(\boldsymbol{X}, \boldsymbol{A}, \boldsymbol{E})$, where $\boldsymbol{A}\in\{0,1\}^{m\times m}$ is the adjacency matrix,  $\boldsymbol{X}\in\mathbb{R}{+}^{m\times d}$ is the node attribute (atom type) matrix ($d$ is the dimension of the node attributes), and $\boldsymbol{E}\in\mathbb{R}{+}^{(m(m-1)/2)\times s}$ denotes the edge attribute (bond type) matrix where $s$ is the number of edge attributes. Note that the real-world molecules have 3-D structures; we leave the GCE for 3-D molecule graphs for future work. 
Furthermore, to determine if a generated counterfactual is classified as desired, we assume that there exists a ground-truth Graph Neural Network (GT-GNN) represented as 
$\phi: \mathcal{D} \rightarrow \mathcal{Y}$, which provides label predictions for graphs in the input graph domain $\mathcal{D}$. 
 This assumption is widely adopted in current literature~\cite{ma2022clear, mahajan2019preserving}.
We define the problem of GCE:
\begin{definition}
    \textbf{\textit{(Graph Counterfactual Explanation)}}. Let $\phi: \mathcal{D} \to \mathcal{Y}$ be the GT-GNN, and let $G=(\boldsymbol{X}, \boldsymbol{A}, \boldsymbol{E})$ be an input graph with $\phi(G) = 0$. The aim of graph counterfactual explanation is to find a model $f: \mathcal D \to \mathcal D$ which computes $f(G) = \hat{G}$ where $\hat{G}$ is a minimally perturbed version of $G$ such that $\phi(\hat{G}) = 1$.
\end{definition}
Here, perturbations on the input graph $G$ may include node/edge insertions and removals as well as changes to node/edge features. We refer to $\hat{G}$ as the counterfactual of the original graph $G$. For an input graph dataset $\mathcal{G}$ sampled from the input graph domain $\mathcal{D}$, we evaluate the performance of a GCE model with the following metrics:
\textbf{\textit{(i)}}\underline{\textit{ \textbf{Validity.}}} Validity measures the fraction of the generated counterfactual graphs $f(G)$ for which $\phi(G) = 0$ and \\$\phi(f(G)) = 1$. Intuitively, it measures how many generated counterfactual graphs actually flip the GT-GNN's prediction of the non-perturbed graph. Accordingly, for convenience, let Valid($\mathcal{G}$) to denote the set $\{f(G) \mid \phi(G) = 0, \phi(f(G)) = 1\}$.  We then define the validity metric of $\mathcal{G}$ as
    \begin{equation}
    \label{eq:validity}
        \textbf{Validity}(\mathcal{G}) = \frac{|\text{Valid}(\mathcal{G})|}{\{G\in\mathcal{G} \mid \phi(G)=0\}}
    \end{equation}
  \textbf{\textit{(ii)}} \underline{\textit{\textbf{Proximity.}}} Proximity measures the mean graph distance $d(\cdot|\cdot)$  between the original graphs $G$ and their generated valid counterfactuals $f(G) = \hat{G}$ (see Appendix~\ref{subapp: details_of_globalGCE}). Low proximity indicates higher-quality counterfactual graphs since they should be made as similar as possible to the graphs they are explaining. The proximity of $\mathcal{G}$ is
    \begin{equation}
        \textbf{Proximity}(\mathcal{G}) = \frac{\sum_{\hat{G} \in \text{Valid}(\mathcal{G})}d(G, \hat{G})}{|\text{Valid}(\mathcal{G})|}.
    \end{equation}
\begin{figure}
\centering
\includegraphics[width=0.45\textwidth]{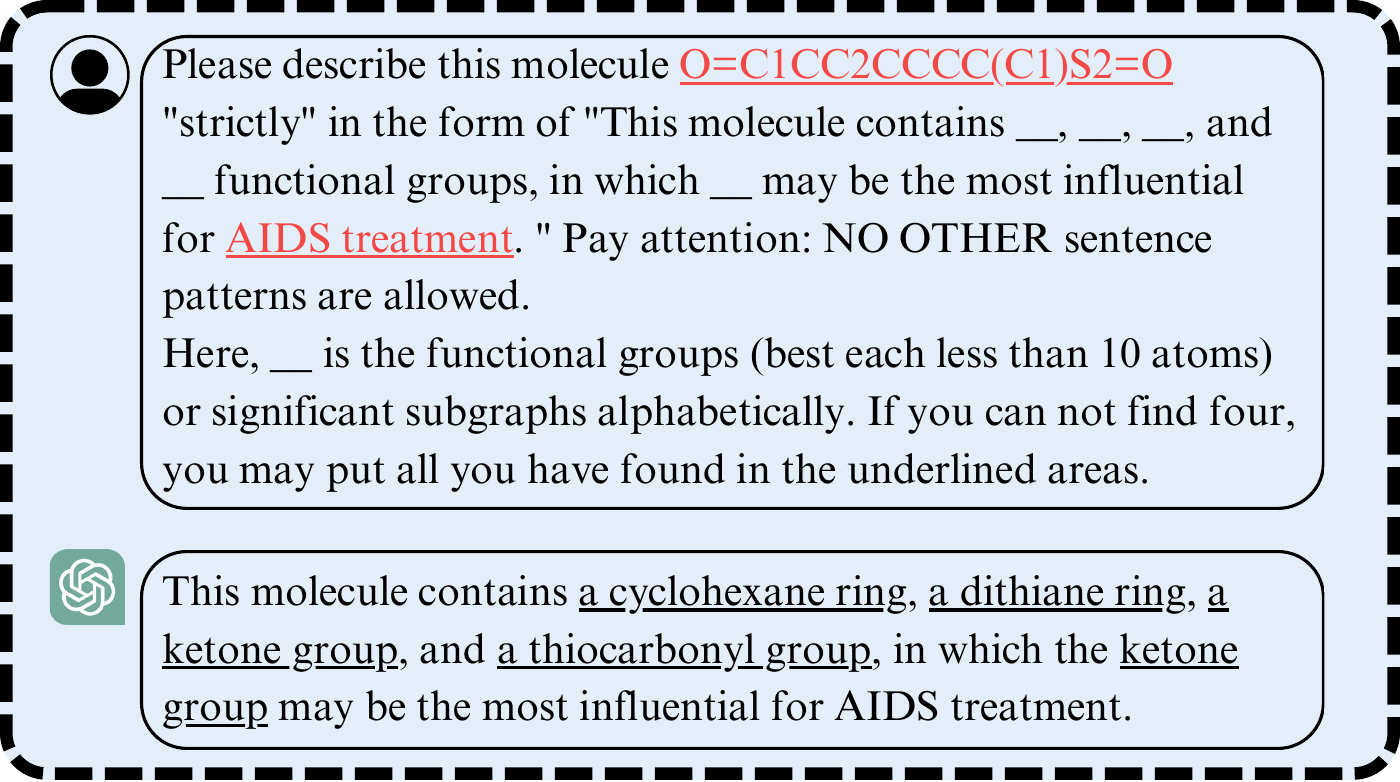}
\vspace{-1mm}
\caption{Prompt-answer pair in text pairs generation.}
\vspace{-0.12in}
\label{fig:TP_sample}
\end{figure}
\begin{figure*}
  \centering
  \vspace{-1mm}
  \includegraphics[width=0.90\textwidth]{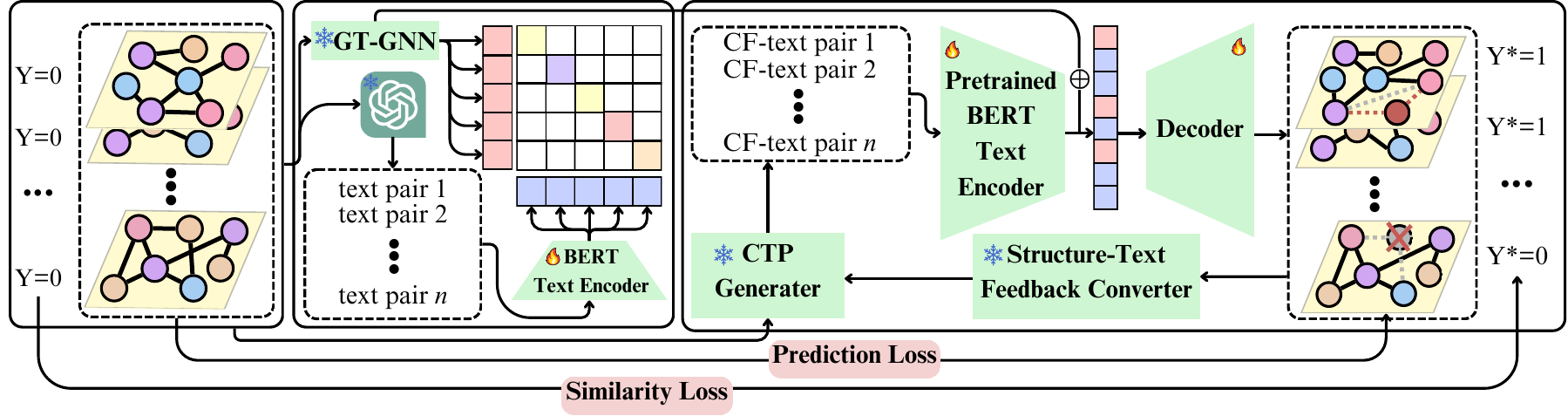}
  \vspace{-3mm}
  \caption{An overview of the proposed LLM-GCE model. 
  }
  \vspace{-15pt}
  \label{fig:overview}
\end{figure*}
In alignment with this, we also provide the validity and proximity results with the \emph{feasibility check}, where we only calculate the two metrics to the set of \emph{feasible} counterfactuals $\hat{G}$s, i.e., those counterfactuals that are chemically stable according to valence-bond theory~\cite{lewis1933chemical}.

%% file: Sections/3-data_preparation.tex
 
There exist multiple datasets of molecules paired with text descriptions~\cite{qian2023can, fang2023mol, zeng2023interactive}; however, most of these datasets only have text pairs describing the graph labeling information.
To support the evaluation of GCE methods, a text pair should contain at least three aspects: graph structure information, graph label information, and the significant subgraphs that contribute the most to the graph's label. Generating satisfying text pairs for GCE is expensive since it requires the most advanced LLMs, such as GPT-4 and customized prompts, so we release five molecule datasets with our generated high-quality text pairs for the convenience of the community.

\noindent\textbf{Construction Process.}
We construct five new text-paired graph datasets based on datasets commonly used in graph explanation~\cite{abrate2021counterfactual, ying2019gnnexplainer, huang2023global}, including AIDS and Mutagenicity from TUDataset~\cite{Morris+2020}; BBBP, ClinTox, and Tox21 from MoleculeNet~\cite{Ramsundar-et-al-2019}. In all datasets, each graph is a molecule with a binary label indicating a molecular property such as AIDS treatment effectiveness (see details in Appendix~\ref{subsubapp:datasets}). 
To generate the text-paired graphs, we take the following steps: (1) \textit{Dataset Preprocessing.} We first convert all input molecular graphs to their SMILES representations~\cite{weininger1988smiles}. We remove molecules that only have one atom or greater than 100 atoms. For Tox21, since the graph label distribution is heavily skewed (more than 95\% of the labels are 0), we randomly select 600 zero-labeled graphs from the dataset. (2) \textit{Text Pair Generation.} Using a custom prompt, we prompt GPT-4 with the SMILES representations and graph labeling semantics for each input graph and ask for a \emph{description} of the molecule's graph structure, graph label, and subgraphs that are most responsible for its label.
(3) \textit{Data Post Processing.} For some graphs, the responses from GPT-4 would be erroneous (i.e., it does not identify the significant subgraphs/functional groups correctly). We fix this by reprompting until desired response is generated.

\noindent \textbf{Prompt Design.}
We generate a text pair (TP) for each graph in a dataset with LLMs, incorporating the graph structure, label semantics, and significant subgraphs. Our prompts use the following template: ``Please describe this graph strictly in the form of `This graph contains \_\_, \_\_, \_\_, and \_\_ significant subgraphs, in which \_\_ may be the most influential for \texttt{the\_label\_semantic}.' No other sentence patterns are allowed. If you can not find four, you may put all you have found in the underlined areas.'' The first half lists the significant subgraphs, revealing the graph's structural information. The second half provides the label semantics so the LLM determines the most significant subgraph for graph labeling.

%% file: Sections/4-method.tex

\subsection{Model Overview}\label{subsec:model_overview}
 An overview of the LLM-GCE model is shown in Fig.~\ref{fig:overview}. The proposed LLM-GCE has three modules: (1) \textit{Contrastive Pretraining of Text Encoder.} We pretrain the text encoder with contrastive learning to align the embeddings of the GT-GNN and the text encoder. (2) \textit{Training of the Counterfactual Autoencoder.} We design a counterfactual autoencoder composed of the pretrained text-encoder and a graph decoder, which is trained to recover the counterfactual topology.
(3) \textit{Dynamic Feedback of CTP Generation}. To tackle hallucination, we prompt the generated counterfactuals with the GT-GNN predictions back to the LLM as the dynamic feedback for further calibration.

\vspace{-4pt}
\subsection{Contrastive Text Encoder Pretraining}\label{subsec:contrastive_pretraining}
\vspace{-2pt}
The first step of LLM-GCE is to pretrain the text encoder so that every TP's embedding aligns with its corresponding graph embedding produced by GT-GNN (via a projection implemented by a multi-layer perception (MLP)~\cite{haykin1994neural}). We choose BERT~\cite{kenton2019bert} as the text encoder due to its proven effectiveness in generating high-quality embeddings for natural language processing tasks. In the graph domain $\mathcal{D}$, we employ a contrastive learning strategy to align the text encoder's embeddings with the GT-GNN's embeddings. First, we sample from $\mathcal{D}$ a dataset $\mathcal{G}=\{G_{i}\}_{i=1}^{n}$ and their TPs. We then train the text encoder with batches of size $N$, which includes $N\times N$ possible graph-TP pairings. Next, we train an MLP to project the TP embeddings from the BERT~\cite{kenton2019bert} text encoder into the embedding space of the GT-GNN and maximize the cosine similarity for the matching graph-TP embedding pairs while minimizing similarity for non-matching pairs.
The contrastive loss is the symmetric cross-entropy given in Appendix~\ref{subapp: contra_loss}.

\subsection{Training of the CA}
After pretraining the text encoder, we generate counterfactuals from natural language. We first provide an overview of our counterfactual autoencoder's (CA) architecture. Then, we elaborate on each of its components, including CTP generation, the text encoder, the latent embedding combination, the graph decoder, and finally, we introduce the overall objective function for our LLM-GCE. 

\subsubsection{\textbf{Architecture Overview.}}\label{subsubsec:architecture_overview} We first generate a CTP, $\text{CTP}_{G}$, for each input graph $G$ as a high-level instruction for GCE. Next, inspired by graph variational autoencoders (VGAEs)~\citep{simonovsky2018graphvae}, we design a CA with a text encoder and a graph decoder. The text encoder, pretrained as introduced in Section~\ref{subsec:contrastive_pretraining}, maps the $\text{CTP}_{G}$ to a probability distribution over a latent space, which is then decoded via an MLP to the counterfactual graph's adjacency matrix $\hat{\boldsymbol{A}}$ and node \& edge attribute matrices $\hat{\boldsymbol{X}}$ \& $\hat{\boldsymbol{E}}$. Training the CA maximizes the likelihood of $\hat{G}$ being a real counterfactual, i.e., $P(\hat{G}|G, \text{CTP}_G, Y^*=1)$.

\subsubsection{\textbf{CTP Generation.}}\label{subsubsection: CTP_generation}
\begin{figure}[!ht]
\centering
\includegraphics[width=0.45\textwidth]{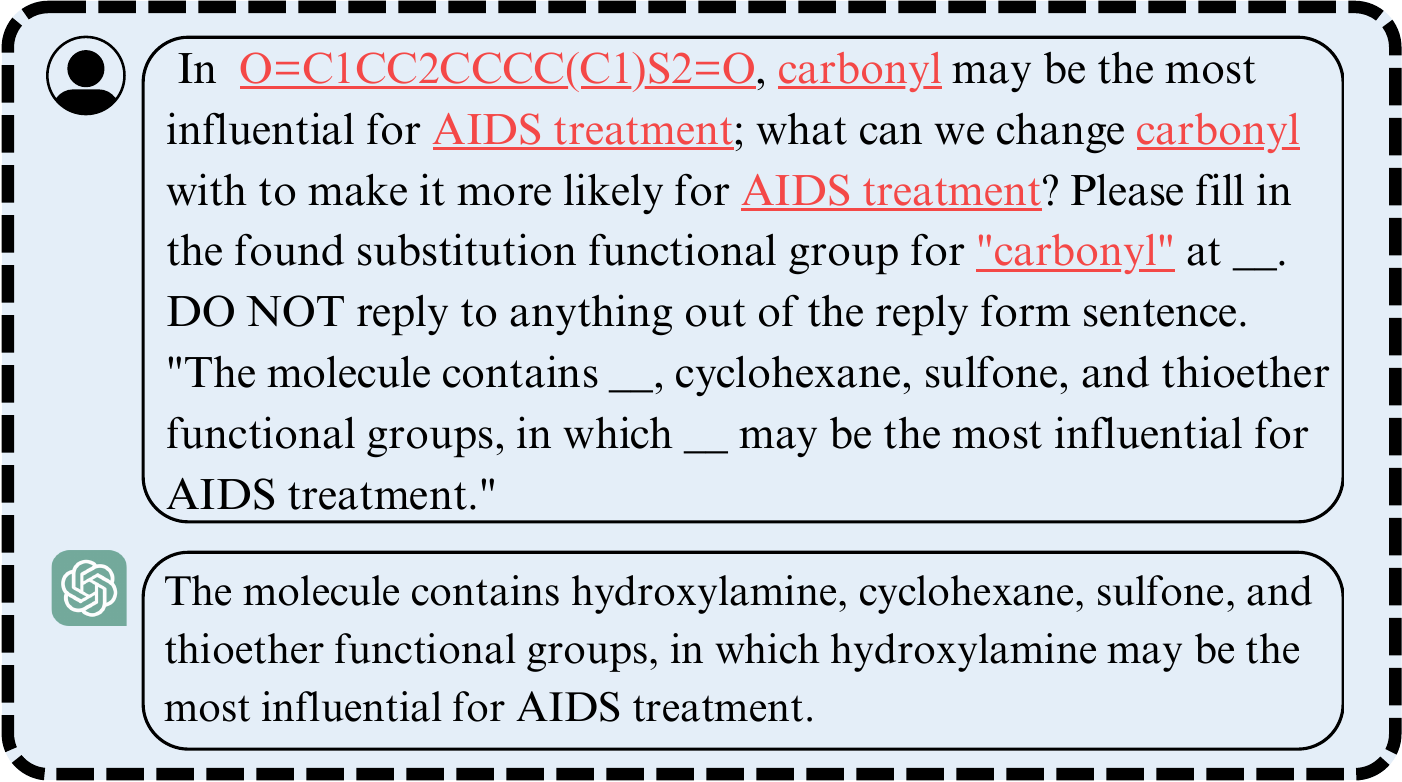}
\vspace{-2mm}
\caption{Illustration of the prompt for CTP generation. The red underlined phrases vary from graph to graph. The LLM strictly follows the input sentence pattern in the answer, which is the same as the one used for TPs.}
\vspace{-0.12in}
\label{fig:CTP_sample}
\end{figure}

For each graph $G$ from the dataset $\mathcal{G}$, we query the LLM for a CTP, a natural language sentence describing the potential counterfactual $\hat{G}$ of $G$. We use this CTP to instruct the generation of counterfactuals from the autoencoder. The CTP generation prompt is shown in Figure~\ref{fig:CTP_sample}, which aims to perform functional group substitution to achieve a higher probability of generating a counterfactual.


\subsubsection{\textbf{Text Embedding Generation.}} The text encoder maps our generated CTPs into a latent space for counterfactual structure reconstruction. 
The input to our text encoder is the CTP of each input graph, denoted as $\{{{\text{CTP}}_{G}}_i\}_{\{i=1\}}^{n}$, where $n = |\mathcal{G}|$ is the number of input graphs. Given $\text{CTP}_G$, the text encoder generates a Gaussian distribution $\mathcal{N}(\boldsymbol{\mu}_G, \boldsymbol{\sigma}_G)$ in the latent space, where $\boldsymbol{\mu}_G\in \mathbb{R}^{l}$ and $\boldsymbol{\sigma}_G\in\mathbb{R}^{l}$ are the distribution's mean and variance. The output of the encoder is a latent embedding $\boldsymbol{z}\in\mathbb{R}^{l}$ sampled from $\mathcal{N}(\boldsymbol{\mu}_G, \boldsymbol{\sigma}_G)$. In practice, the text encoder is implemented with BERT~\cite{kenton2019bert} and initialized with the pretrained parameters from Section~\ref{subsec:contrastive_pretraining}. Recall that our contrastive pretraining objective encourages the latent text embedding of each $\text{TP}_G$
to be close to its corresponding graph latent embedding provided by the GT-GNN under a projection by an MLP. Therefore, although the exact graph structure, node, and edge attributes of the desired counterfactual $\hat{G}$ are not immediately accessible, the embedding of the CTP 
approximates $\hat{G}$'s GT-GNN embedding, which guides counterfactual decoding. Intuitively, the text encoder's embeddings contain the information of a high-level counterfactual generation instruction, while the GT-GNN's embeddings encode each counterfactual's graph structure and node \& edge attributes. Next, we introduce how we combine the embedding of the text encoder with that of the GT-GNN.
\subsubsection{\textbf{Latent Embedding Combination.}}
The generated CTP and acquired latent embedding from the text encoder for an input graph $G$ are still insufficient for counterfactual generation. This is for two reasons. (1) Each CTP, as introduced in Section~\ref{subsubsection: CTP_generation}, only contains information about the significant subgraphs (e.g., functional groups for molecule graphs) of the counterfactual $\hat{G}$ while the specific structure of the counterfactual is not described in detail. (2) While the pretraining process enhances the consistency of each text embedding with its corresponding GT-GNN embedding, the limited availability of a large amount of pretraining data prevents the model from achieving high pretraining accuracy. Therefore, to merge information from both the counterfactual text and graph embeddings, we update the final encoded embedding $e_G$ by concatenating the text encoder embeddings $\boldsymbol{z}_{G}$ of $\{{\text{CTP}_{G}}_i\}_{\{i=1\}}^{n}$ with the GT-GNN embeddings $\boldsymbol{q}_{G}$ of the input graphs $\mathcal{G}$, written as $\boldsymbol{e}_{G}=\boldsymbol{z}_{G}\oplus\boldsymbol{q}_{G}$. 
\subsubsection{\textbf{Graph Decoder.}} 
In the graph decoder, the resulting latent embeddings ${{\{\boldsymbol{e}_{G}}_{i}\}}_{i=1}^{n}$ are used to reconstruct the adjacency matrix $\hat{\boldsymbol{A}}_{G}\in \mathbb{R}^{m\times m}$, node attribute matrix $\hat{\boldsymbol{X}}_{G}\in \mathbb{R}^{m\times d}$, and edge attribute matrix $\hat{\boldsymbol{E}}_{G}\in \mathbb{R}^{m(m-1)/2\times s}$ of the counterfactual graph $\hat{G}$ for each input graph $G$. The graph decoder is implemented as an MLP with a sigmoid activation, restricting its output range to $(0, 1)$. As a result, every entry of the generated matrices is a continuous probabilistic real number.
However, a real graph's adjacency matrix is a ${0, 1}$-matrix, where $\hat{\boldsymbol{A}}_{ij}=1$ indicates the presence of an edge between nodes $i$ and $j$, and $\hat{\boldsymbol{A}}_{ij}=0$ indicates the absence of an edge. Furthermore, each row of the node/edge attribute matrix is a one-hot code indicating the discrete node/edge type. To make the decoder compatible with these constraints, we discretize the generated adjacency matrix $\hat{\boldsymbol{A}}_{\hat{G}}$ by thresholding its probabilistic entries, setting entries to 1 if the corresponding value exceeds 0.5 and to 0 otherwise. Similarly, we generate the one-hot node and edge attribute matrices $\hat{\boldsymbol{X}}_{\hat{G}}$ and $\hat{\boldsymbol{E}}_{\hat{G}}$ by taking the one-hot row-wise argmax of each probabilistic matrix.
\subsubsection{\textbf{Objective Function.}}
As discussed in Section~\ref{subsubsec:architecture_overview}, the optimization target is to maximize the likelihood of the generated graph being a real counterfactual $\hat{G}$ conditioned on $\text{CTP}_G$ and the desired label $Y^*$, denoted as $P(\hat{G}|G, \text{CTP}_{G}, Y^*=1)$. We formalize this objective with the Kullback-Leibler (KL) divergence~\cite{csiszar1975divergence} of the distribution given by the encoder $Q(\boldsymbol{e}|G, \text{CTP}_{G}, Y^*)$ and the posterior distribution $P(\boldsymbol{e}|\hat{G}, \text{CTP}_{G}, Y^*)$. For simplicity, we write the condition $\{\text{CTP}_{G}, Y^*\}$ as $T$, and the divergence term $\text{KL}[Q(\boldsymbol{e}|G, T)||P(\boldsymbol{e}|\hat{G}, T)]
    =-\mathbb{E}_{\boldsymbol{e}\sim Q}[\log{P(\boldsymbol{e}|\hat{G},T)
    -\log{Q(\boldsymbol{e}|G, T)}}]$.
By $\log{P(\boldsymbol{e}|\hat{G},T)}=\log{P(\boldsymbol{e}|T)}+\log{P(\hat{G}|\boldsymbol{e}, T)}$, we have the equation 
\begin{equation}
\begin{aligned}
&\log{P(\hat{G}|\text{CTP}_{G}, Y^*)}-
    \text{KL}[Q||P] \\
    =&\mathbb{E}_{\boldsymbol{e}\sim Q}[\log{P(\hat{G}|z, T)}]
    -\text{KL}[Q||P],
\end{aligned}
\end{equation}
where $Q=Q(\boldsymbol{e}|G, T)$ and $P=P(\boldsymbol{e}|\hat{G}, T)$. In this equation, the first term in the left-hand side (LHS) is our optimization target, and the second term is a KL divergence, which is inaccessible since the posterior $P(\boldsymbol{e}|\hat{G})$ is intractable. However, the right-hand side is available for direct calculation. Therefore, we optimize the log-likelihood $P(\hat{G}|G, \text{CTP}_{G}, Y^*=1)$ with its Evidence Lower Bound (ELBO)~\cite{kingma2013auto}:
\begin{equation}\label{equ:elbo_geq}
\begin{aligned}
    \log{P(\hat{G}|T)}
    \geq&\mathbb{E}_{\boldsymbol{e}\sim Q}[\log{P(\hat{G}|z,T)}] \\
    &-\text{KL}[Q(\boldsymbol{e}|G, T)||P(\boldsymbol{e}|T)].
\end{aligned}
\end{equation}
 However, due to the lack of the ground-truth counterfactual $\hat{G}$, we substitute the first term of RHS in Equ.~(\ref{equ:elbo_geq}) with two loss terms. (1) \emph{Graph Distance Loss ($\mathcal{L}_{\text{dist}}$)} encourages small graph distances between $G$ and its counterfactual $\hat{G}$. Formally,  
 \begin{equation}
 \mathcal{L}_{\text{dist}}:=\Sigma_{\{G\in \mathcal{G}\}}r(G,\hat{G}),
 \end{equation} 
 where $r(\cdot, \cdot)$ is the weighted sum of the distances between graph adjacency matrices $r_{\boldsymbol{A}}(\cdot, \cdot)$, node attribute matrices $r_{\boldsymbol{X}}(\cdot, \cdot)$, and edge attribute matrices $r_{\boldsymbol{E}}(\cdot, \cdot)$, i.e., $r(G,\hat{G})=r_{\boldsymbol{A}}(\boldsymbol{A}_{G}, \boldsymbol{A}_{\hat{G}})+r_{\boldsymbol{X}}(\boldsymbol{X}_{G}, \boldsymbol{X}_{\hat{G}})+r_{\boldsymbol{E}}(\boldsymbol{E}_{G}, \boldsymbol{E}_{\hat{G}})$ (see Appendix~\ref{subapp: details_of_globalGCE} for definitions). (2) \emph{Counterfactual Prediction Loss ($\mathcal{L}_{\text{pred}}$)} is the log-likelihood that the generated counterfactual is classified as desired by GT-GNN:
 \begin{equation}
     \mathcal{L}_{\text{pred}}:= -\log{P_{\text{GT-GNN}}(\phi(\hat{G})=Y^*)}.
 \end{equation}
In conclusion, our overall loss function is 
\begin{equation}
\begin{aligned}
    \mathcal{L} =& \alpha \mathcal{L}_{\text{dist}}+\beta \mathcal{L}_{\text{pred}}- \mathcal{L}_{\text{KL}},
\end{aligned}
\end{equation}
$\mathcal{L}_{\text{KL}}=\text{KL}[Q(\boldsymbol{e}|G, \text{CTP}_{G}, Y^*)||P(\boldsymbol{e}|\text{CTP}_{G}, Y^*)$.

\subsection{Dynamic Feedback of CTP Generation}
\begin{figure}[!ht]
\centering
\includegraphics[width=0.45\textwidth]{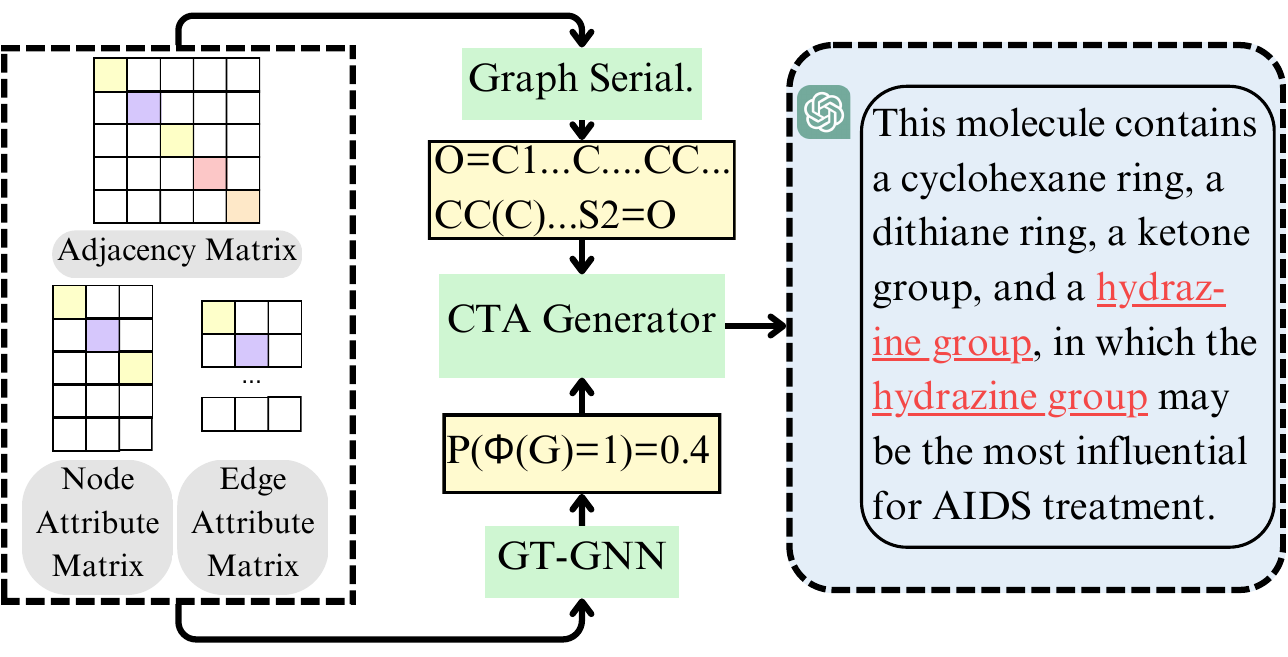}
\vspace{2mm}
\caption{
Illustration of the CTP's dynamic feedback. ``Graph Serial.'' is short for ``Graph serialization'' which converts graphs to their SMILES representations. 
}
\label{fig:dynamical_feedback}
\end{figure}
The CTP generator acts as a commander, giving high-level instructions (CTPs) for counterfactual graph generation. The CA is the executor, implementing the instruction to create a specific counterfactual graph. However, CTPs can be inaccurate due to LLM hallucination~\cite{zhang2023siren} and limited graph decoding ability. To remedy this, a CTP dynamic feedback scheme is designed to calibrate CTP with the CA-generated counterfactuals.

An illustration of the scheme is shown in Fig.~\ref{fig:dynamical_feedback}. We first convert the counterfactual molecule generated by the CA into its SMILES representation. Then, we combine it with the GT-GNN's probability that the graph is a valid counterfactual into a specific prompt asking for a new, calibrated CTP for the original graph $G$. This concludes a single iteration of dynamic feedback, which we treat as a hyperparameter. Each iteration of dynamic feedback can be seen as an indirect reasoning step, forcing the model to reflect on its past outputs and the label information from the GT-GNN in order to produce more truthful CTPs. A similar calibration approach is verified effective in~\citet{dhuliawala2023chain,madaan2024self}


%% file: Sections/5-experiments.tex
In this section, we evaluate LLM-GCE with extensive experiments on five real-world datasets. Our experiments aim to answer the following research questions (RQs): \textbf{RQ1}: How does LLM-GCE perform w.r.t. validity and proximity compared with state-of-the-art baselines? \textbf{RQ2}: How does each component of LLM-GCE affect its overall performance? \textbf{RQ3:} What insights can LLM-GCE provide given its counterfactual explanation results?

\begin{table*}
\centering
\setlength{\tabcolsep}{3pt}
\renewcommand{\arraystretch}{0.96}
\caption{The performance of different GCE methods. The best results are in bold, and the runner-up results are underlined. `n/a' refers to unavailable proximity scores since there is no valid counterfactual graph.}
\vspace{-2.5mm}
\label{tab:main_table}
\resizebox{\textwidth}{!}{\begin{tabular}{c|c|c|ccccc}
\hline
                              &                            &           & \textbf{GNNExplainer}                 & \textbf{CF-GNNExplainer}              & \textbf{CLEAR}                         & \textbf{RegExplainer}                 & \textbf{LLM-GCE}                \\ \hline
\multirow{4}{*}{\textbf{AIDS}}         & \multirow{2}{*}{\textbf{Validity}}  & w. Feas.  & $\underline{0.25 \pm 0.00}$  & $\underline{0.25 \pm 0.00}$  & $\bm{2.56 \pm 3.10}$          & $\underline{0.25 \pm 0.00}$  & $0.25 \pm 0.27$        \\
                              &                            & w/o Feas. & $\bm{100.00 \pm 0.00}$       & $\bm{100.00 \pm 0.00}$       & $\bm{100.00 \pm 0.00}$        & $\bm{100.00 \pm 0.00}$       & $\bm{100.00 \pm 0.00}$      \\ \cline{2-8} 
                              & \multirow{2}{*}{\textbf{Proximity}} & w. Feas.  & $\bm{2.00 \pm 0.00}$         & n/a                          & $109.82 \pm 1.18$             & $7.28 \pm 0.00$  & $\underline{5.86 \pm 5.62}$        \\
                              &                            & w/o Feas. & $\bm{4.27 \pm 0.00}$         & n/a                          & $109.57 \pm 0.96$             & $\underline{15.11 \pm 0.00}$ & $86.37 \pm 2.65$     \\ \hline
\multirow{4}{*}{\textbf{Mutagenicity}} & \multirow{2}{*}{\textbf{Validity}}  & w. Feas.  & $0.00 \pm 0.00$              & $0.00 \pm 0.00$              & $\underline{4.89 \pm 6.92}$   & $0.00 \pm 0.00$              & $\bm{13.52 \pm 9.93}$  \\
                              &                            & w/o Feas. & $\underline{65.96 \pm 0.00}$ & $50.17 \pm 2.32$             & $\bm{100.00 \pm 0.00}$        & $\bm{100.00 \pm 0.00}$       & $\bm{100.00 \pm 0.00}$ \\ \cline{2-8} 
                              & \multirow{2}{*}{\textbf{Proximity}} & w. Feas.  & n/a                          & n/a                          & $\underline{406.42 \pm 0.00}$ & n/a                          & $\bm{1.75 \pm 2.48}$  \\
                              &                            & w/o Feas. & $27.71 \pm 0.078$            & $\bm{8.77 \pm 0.48}$         & $409.52 \pm 1.11$             & $\underline{24.91 \pm 0.00}$ & $56.31 \pm 2.57$     \\ \hline
\multirow{4}{*}{\textbf{BBBP}}         & \multirow{2}{*}{\textbf{Validity}}  & w. Feas.  & $0.74 \pm 0.00$              & $0.00 \pm 0.00$              & $\underline{9.56 \pm 8.34}$   & $0.74 \pm 0.00$              & $\bm{38.25 \pm 10.25}$     \\
                              &                            & w/o Feas. & ${22.79 \pm 0.00}$           & $\underline{34.38 \pm 4.42}$ & $\bm{100.00 \pm 0.00}$        & $\bm{100.00 \pm 0.00}$       & $\bm{100.00 \pm 0.00}$ \\ \cline{2-8} 
                              & \multirow{2}{*}{\textbf{Proximity}} & w. Feas.  & $\underline{16.06 \pm 0.00}$        & n/a                          & $141.85 \pm 0.71$             & $18.25 \pm 0.00$ & $\bm{11.68 \pm 0.14}$         \\
                              &                            & w/o Feas. & $\underline{17.26 \pm 0.00}$ & $\bm{6.84 \pm 0.76}$         & $142.00 \pm 0.11$             & $27.45 \pm 0.00$             & $55.98 \pm 5.79$       \\ \hline
\multirow{4}{*}{\textbf{ClinTox}}      & \multirow{2}{*}{\textbf{Validity}}  & w. Feas.  & $0.00 \pm 0.00$              & $0.00 \pm 0.00$              & $\underline{1.33 \pm 1.89}$   & $0.00 \pm 0.00$              & $\bm{30.86 \pm 26.41}$     \\
                              &                            & w/o Feas. & $\bm{100.00 \pm 0.00}$       & $\bm{100.00 \pm 0.00}$       & $\bm{100.00 \pm 0.00}$        & $\bm{100.00 \pm 0.00}$       & $\bm{100.00 \pm 0.00}$ \\ \cline{2-8} 
                              & \multirow{2}{*}{\textbf{Proximity}} & w. Feas.  & n/a                          & n/a                          & $\underline{50.10 \pm 72.27}$ & n/a                          & $\bm{0.95 \pm  0.93}$   \\
                              &                            & w/o Feas. & $\bm{16.61 \pm 0.00}$        & n/a                          & $150.50 \pm 0.97$             & $\underline{28.74 \pm 0.00}$ & $42.89 \pm 3.72$     \\ \hline
\multirow{4}{*}{\textbf{Tox21}}        & \multirow{2}{*}{\textbf{Validity}}  & w. Feas.  & $0.00 \pm 0.00$              & $0.00 \pm 0.00$              & $0.00 \pm 0.00$               & $0.00 \pm 0.00$              & $\bm{3.70 \pm 4.01}$   \\
                              &                            & w/o Feas. & $\underline{0.21 \pm 0.00}$  & ${0.00 \pm 0.00}$            & $\bm{100.00 \pm 0.00}$        & ${0.00 \pm 0.00}$            & $\bm{100.00 \pm 0.00}$      \\ \cline{2-8} 
                              & \multirow{2}{*}{\textbf{Proximity}} & w. Feas.  & n/a                          & n/a                          & n/a                           & n/a                          & $\bm{5.93 \pm 4.91}$  \\
                              &                            & w/o Feas. & $\bm{18.28 \pm 0.00}$        & n/a                          & $190.70 \pm 1.72$ & n/a                          & $\underline{77.34 \pm 4.28}$   \\ \hline
\end{tabular}}
\end{table*}
\subsection{Experimental Setup}
\subsubsection{\textbf{Datasets.}}
Our experiments utilize five real-world datasets (AIDS, Mutagenicity, BBBP, ClinTox, Tox21). Among them, AIDS and Mutagenicity are from TUDataset~\cite{Morris+2020}, while BBBP, SIDER and Tox21 are from MoleculeNet~\cite{Ramsundar-et-al-2019}. In these datasets, graphs represent chemical compounds with nodes as atoms and edges as bonds.  They are labeled based on relevance to properties such as blood-brain barrier penetration, mutagenicity, HIV activity, side effects resource, and toxicological activity. We associate each graph with a text pair by the procedure in Section~\ref{sec:data_prepare}. For details, see Appendix~\ref{subsubapp:datasets}.
\subsubsection{\textbf{Baselines}}
We adopt the following state-of-the-art baselines. (1) \textbf{GNNExplainer}~\citep{ying2019gnnexplainer} is a graph factual explanation (GFE) model. GFE is a graph explanation strategy slightly different from GCE, and so we adjust it for GCE by revising its loss function. See Appendix~\ref{subsup: baselines} for details. (2) \textbf{CF-GNNExplainer}~\citep{lucic2022cf} is targeted at node-level GCE. We adapt it for graph-level GCE by changing the input from ego-graphs to whole graphs and changing the supervisory signal from node labels to graph labels. (3) \textbf{CLEAR}~\citep{ma2022clear} is a generative GCE model that is based on graph variational autoencoders~\cite{simonovsky2018graphvae}. We adapt it by removing its causality-specific component for a fair comparison. (4) \textbf{RegExplainer}~\citep{zhang2023regexplainer} is a GCE model for graph regression that can be directly adapted to graph classification. We evaluate those GCE baselines based on Appendix~\ref{subapp: details_of_globalGCE}.  
\subsubsection{\textbf{GNN Classifier}}
Before training LLM-GCE, we first train a two-layer Graph Convolutional Network (GCN) slightly modified to incorporate edge attributes (see Appendix~\ref{subsupp: gnn_prediction_model}) for the five datasets to serve as the GT-GNN. 
The GT-GNN has a node embedding dimension of 32, a maximum pooling layer, and a fully connected layer for graph classification. We set the edge embedding dimension $d$ to 1.
The model is trained with the Adam optimizer~\citep{kingma2014adam} with a learning rate 1e-3 for 500 epochs. The train/validate/test split is 50\%/25\%/25\%. Accuracies are shown in Table~\ref{tab:classifier_accuracy}.
\begin{table}
    \centering
    \setlength{\tabcolsep}{6.9pt}
\renewcommand{\arraystretch}{1.1}
    \caption{GT-GNN accuracy (\%) on adopted datasets.}
    \label{tab:classifier_accuracy}
    \vspace{-5pt}
    \small
\resizebox{\columnwidth}{!}{\begin{tabular}{cccccl}
\hline
\multicolumn{1}{l}{} & \multicolumn{1}{l}{\textbf{AIDS}} & \multicolumn{1}{l}{\textbf{Muta.}} & \multicolumn{1}{l}{\textbf{BBBP}} & \multicolumn{1}{l}{\textbf{ClinTox}} & \multicolumn{1}{l}{\textbf{Tox21}} \\ \hline
\textbf{Training}    &   0.995   &    0.781  &   0.887   &   0.749   &   0.970   \\
\textbf{Validation}  &   0.996   &    0.722  &   0.852   &   0.782   &   0.972   \\
\textbf{Testing}     &   0.994   &    0.762  &   0.850   &   0.783   &   0.974    \\ \hline
\end{tabular}}
\vspace{-12pt}
\end{table}
\vspace{-5pt}
\subsubsection{\textbf{Explainer Settings.}}
We detail the experimental settings for pretraining (the second part in Fig.~\ref{fig:overview}) and training (the third part in Fig.~\ref{fig:overview}) stages of our pipeline. 
The list of all the prompts used are in Appendix~\ref{subsupp: prompts}.

\noindent \textbf{Pretraining.} 
We use GPT-4 as the TP generator. The BERT text encoder is implemented through Huggingface~\cite{wolf-etal-2020-transformers}. We use the AdamW optimizer~\cite{loshchilov2018decoupled} for pretraining BERT over 100 epochs with a learning rate of 0.01. 

\noindent \textbf{Training.} 
We choose GPT-3.5-Turbo as the CTP generator, where feedback is performed three iterations in every epoch. Finetuning hyperparameters are the same as those used in pretraining.
\subsection{RQ1: Performance of Different Methods}
We evaluate our LLM-GCE framework on five real-world datasets and compare its validity and proximity performance against state-of-the-art baselines, with and without a feasibility (Feas.) check (Table~\ref{tab:main_table}) as determined by checking for stability under valence theory with RDKit~\citep{rdkit}. We have the following observations:
(1) From the perspective of validity, LLM-GCE achieves comparable validity with other baselines without the chemical feasibility check. However, with the feasibility check, our model achieves the highest validity among almost all baselines across all datasets.
(2) From the perspective of proximity, LLM-GCE achieves the lowest proximity among chemically feasible counterfactuals among almost all baselines. This means that our model can find valid counterfactuals that are not only feasible but also have a minimal graph distance from their corresponding input graph.
(3) Based on these observations, LLM-GCE achieves satisfying graph counterfactual explanation performance, especially when chemical feasibility is considered. This reveals the effectiveness of LLM's pretrained knowledge and reasoning abilities in GCE.
For details on the efficiency of LLM-GCE, see Appendix~\ref{subapp: model_efficiency}.
\subsection{RQ2: Ablation Study}
\begin{figure}[t]
\centering
\vspace{-1mm}
\includegraphics[width=0.75\columnwidth]{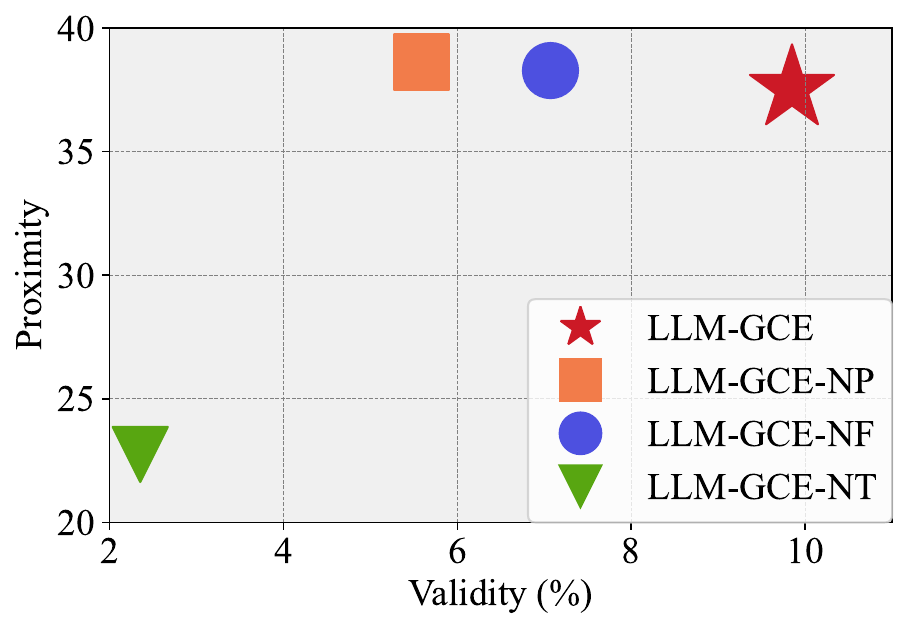}
\vspace{-3mm}
\caption{
Ablation study on BBBP. -NP: No Pretraining. -NF: No Feedback. -NT: Bert Autoencoder Frozen.
}\label{fig:ablation_study}
\vspace{-4mm}
\end{figure}

We experiment with three ablated variants of LLM-GCE:
\textbf{LLM-GCE-NP}: Without pretraining our BERT text-encoder, we directly train the counterfactual autoencoder with a dynamic CTP feedback module. \textbf{LLM-GCE-NT}: We freeze the counterfactual autoencoder in LLM-GCE, i.e., the counterfactual autoencoder is not optimized between two dynamic CTP feedback iterations. \textbf{LLM-GCE-NF}: We remove the CTP dynamic feedback module and generate graph counterfactuals only from the autoencoder and the initial CTP. 
From Fig.~\ref{fig:ablation_study}, we make the following observations.
(1) LLM-GCE-NP: Refraining from pretraining decreases model validity and increases proximity.
The small decline in performance is possibly due to insufficient data for contrastive learning.
(2) LLM-GCE-NT: Validity dramatically degrades when the autoencoder is frozen.
Proximity also decreases, possibly because a frozen autoencoder can only generate a small number of counterfactuals that are less diverse.
(3) LLM-GCE-NF: Removing the feedback module significantly reduces validity and slightly increases proximity, revealing the importance of dynamic feedback for GCE. We have similar observations on other datasets. For more ablation studies, see Appendix~\ref{subapp: para_sensitivity}. Additionally, we substitute the text encoder of the CA and CTP generator to other language models, see the results in Appendix~\ref{subapp: various_LLM}.
\begin{figure}[t]
\centering
\includegraphics[width=0.5\columnwidth]{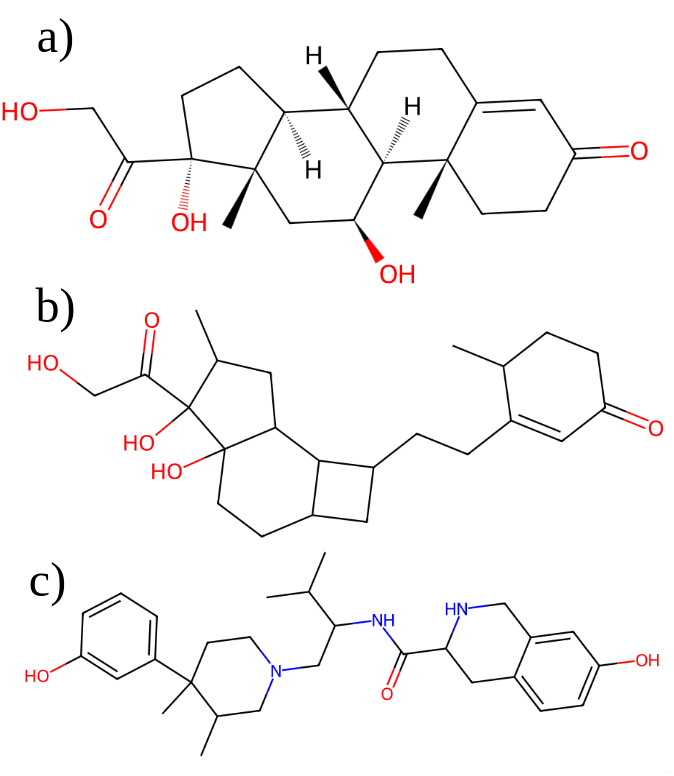}
\caption{
Generated molecules by LLM-GCE and GNNExplainer. a) from ClinTox, b) by LLM-GCE, c) by GNNExplainer. Proximity of b) is 16.52, c) is 24.40.
}\label{fig:proximity-compare}
\vspace{-15pt}
\end{figure}

\subsection{RQ3: Case Study}
We highlight two main advantages of LLM-GCE: (1) \textit{More Faithful Counterfactuals.} The generated counterfactuals by LLM-GCE have consistently lower proximity across datasets than that of the baseline methods. We show a counterfactual generated by our LLM-GCE compared with that by GNNExplainer in Fig. \ref{fig:proximity-compare}. Molecule a) is from the ClinTox dataset, b) is generated by LLM-GCE, and c) by GNN Explainer. LLM-GCE's output better preserves the original molecule's structural integrity than GNNExplainer.
(2) \textit{More Feasible Counterfactuals.} With the LLMs' extensive domain knowledge, our LLM-GCE generates more counterfactuals satisfying valence bond theory, while those given by baselines do not satisfy this theory. Only CLEAR generates a high proportion of feasible ones, but they reveal no chemical insight as most are disconnected carbon atoms, misleadingly passing the feasibility check~\cite{weininger1988smiles}. For more case studies, see Appendix~\ref{subapp: case_studies}.
\begin{figure}[t]
    \centering
    \begin{minipage}{.235\textwidth}
        \centering
        \includegraphics[width=1.0\linewidth]{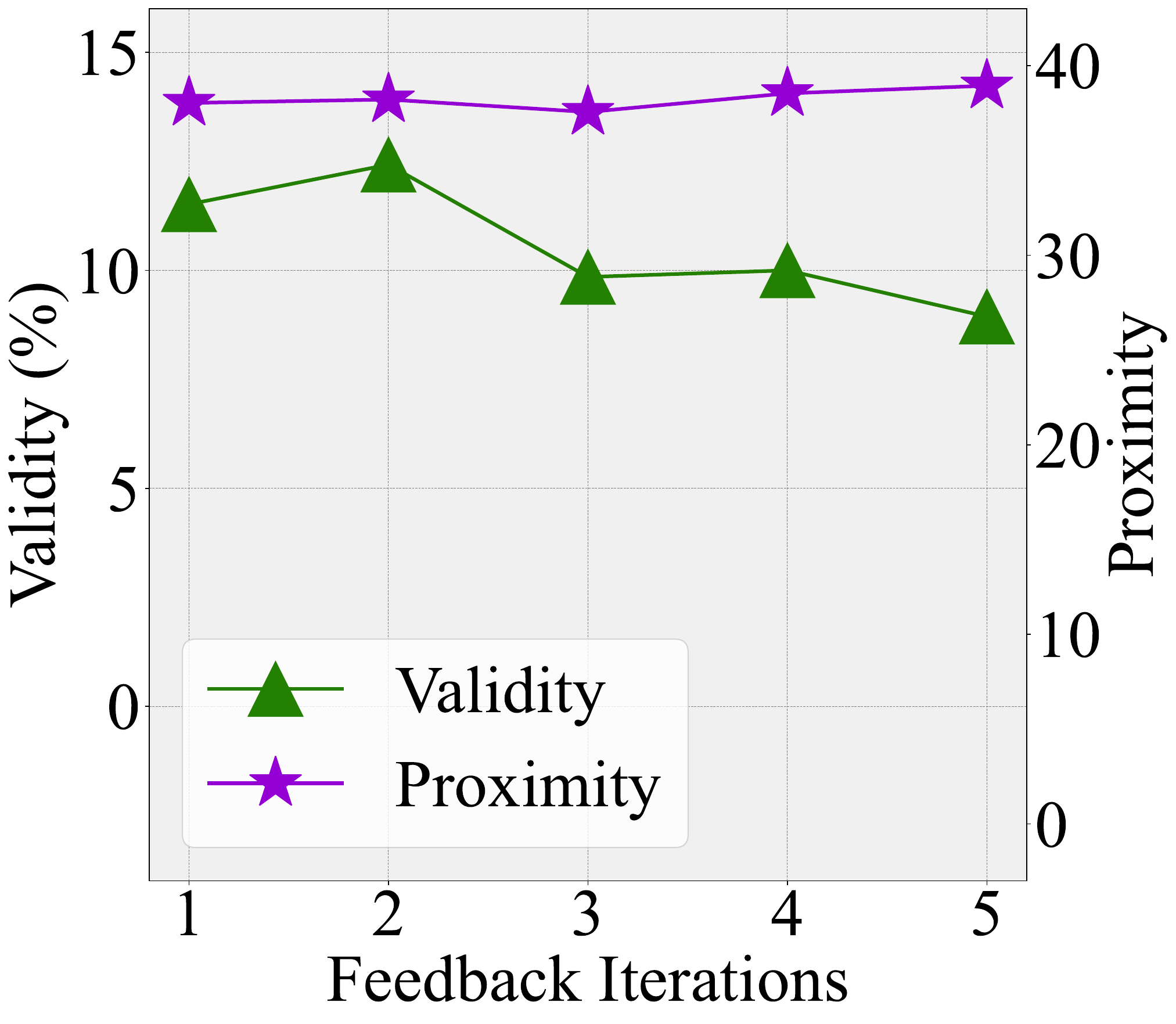}
        \label{fig:feedback_times}
    \end{minipage}
    \hfill 
    \begin{minipage}{.235\textwidth}
        \centering
        \includegraphics[width=1.0\linewidth]{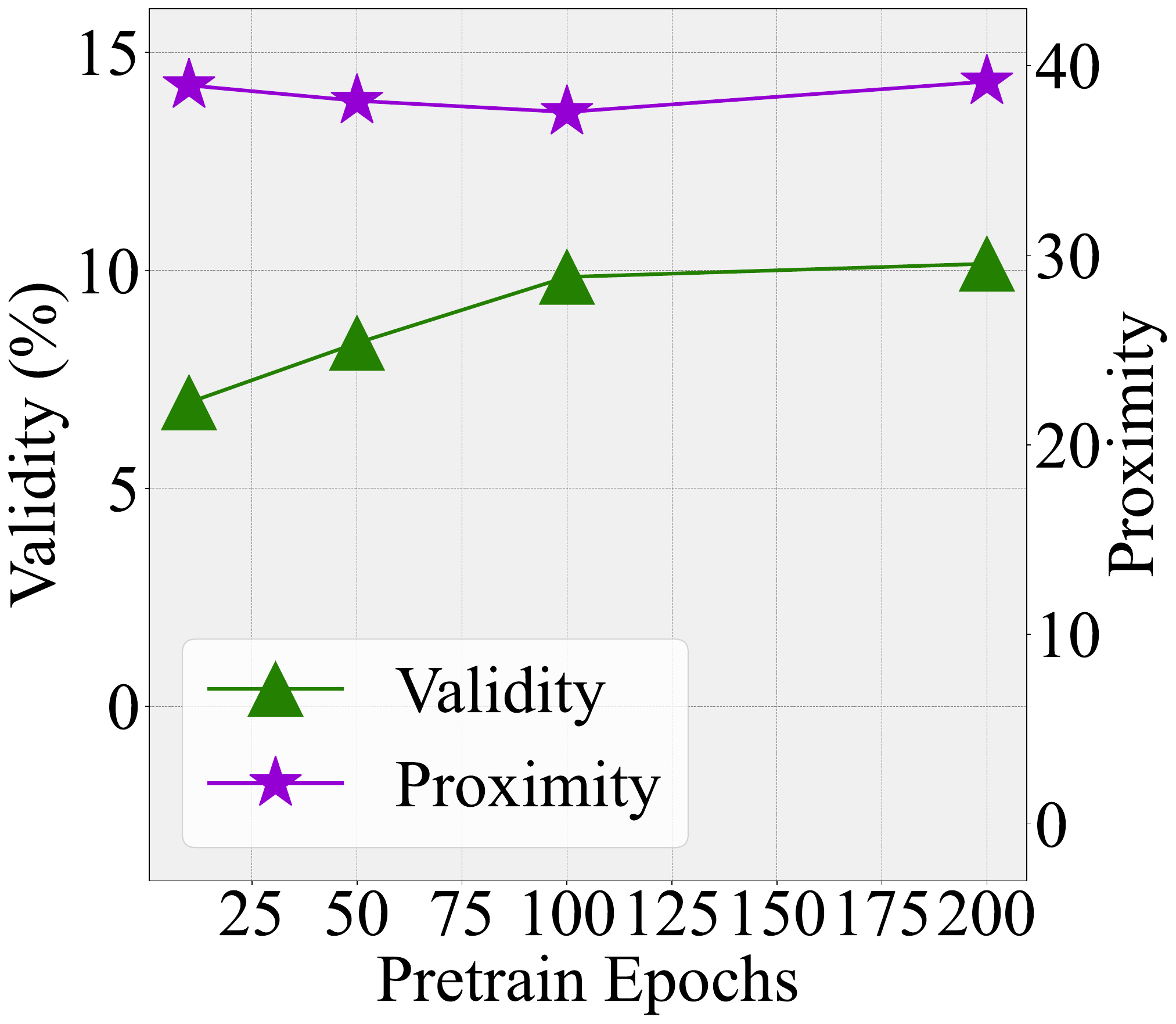}
        \label{fig:pretrain_epochs}
    \end{minipage}
    \vspace{-10pt}
    \caption{Parameter Analysis on BBBP. (a) Validity and proximity w.r.t. CTP feedback iterations. (b) Validity and proximity w.r.t. text encoder pretrain epochs.}
    \label{fig:parameter_study}
    \vspace{-15pt}
\end{figure}
\vspace{-5pt}
\subsection{RQ4: Parameter Analysis}
We study how dynamic CTP feedback iterations and text encoder pretrain epochs affect LLM-GCE's performance on BBBP. We test feedback iterations from 1 to 5 and pretraining epochs in \{10, 50, 100, 200\}. 
We observe from Fig.~\ref{fig:parameter_study} that
(1) validity improvement saturates after two feedback iterations while proximity remains nearly unchanged. While a small number of feedback iterations can enhance performance, it is possible that simply increasing rounds worsens hallucinations.
Also, (2) increasing pretraining epochs to around 100 boosts validity. At the same time, proximity initially improves up to 100 epochs, but then gradually increases. This trend suggests a correlation between validity and proximity regarding pretraining epochs, with the ideal count being approximately 100. Similar findings are observed on other datasets. For more results, see Appendix~\ref{subapp: para_sensitivity}.

%% file: Sections/7-conclusions.tex
In this work, we explore the ability of LLMs in guiding the GCE for molecule properties prediction. Specifically, we propose a novel model called LLM-GCE, comprised of a contrastive pre-training module, a counterfactual autoencoder, and a dynamic feedback module. Extensive experiments validate the superior performance of LLM-GCE.


%% file: Sections/9-acknowledgement.tex
This work is supported in part by the National Science Foundation under grants (IIS-2006844, IIS-2144209, IIS-2223769, CNS-2154962, BCS-2228534, and CMMI-2411248), Office of Naval Research under grant (N000142412636), the Commonwealth Cyber Initiative Awards under grants (VV-1Q24-011, VV-1Q24-011), and the research gift funding from Netflix and Snap.
 

%% file: Sections/8-limitations.tex
Firstly, the effectiveness of LLM-GCE relies heavily on the quality and relevance of the pretraining data used for the large language models. If the pretraining data is biased or lacks sufficient coverage of the target domain, it may lead to less accurate or relevant counterfactuals being generated. Ensuring the LLMs are trained on high-quality, domain-specific data is crucial for optimal performance.

Additionally, the computational cost associated with using large language models can be a drawback. Training and inference with LLMs are more time-consuming and resource-intensive than other graph counterfactual explanation methods. Although we have shown that the time execution time is comparable for our methods with other baselines on the adopted datasets, LLM-GCE can be time consuming when dealing with extremely large-scale graphs, such as some big proteins.

Furthermore, while the experiments demonstrate the effectiveness of LLM-GCE on several real-world datasets, the evaluation is still limited to specific domains, such as molecular property prediction. The generalizability of the proposed framework to other types of graphs and application areas remains to be investigated. Further research is needed to assess the performance and adaptability of LLM-GCE across a wider range of graph structures and problem domains.

Lastly, the potential for hallucinations or inconsistencies in the generated counterfactuals remains a solid challenge. Although the dynamic feedback module aims to mitigate this issue, there may still be cases where the LLMs produce counterfactuals that are not entirely faithful to the original graph or the desired properties. 


%% file: Sections/Ethics.tex
In this work, we propose a novel LLM-guided graph counterfactual explanation method that utilizes the strong reasoning ability of LLMs. 
We do not anticipate any ethical issues that should be specifically highlighted in this paper.

%% file: Appendix/supplementary_material.tex
\appendix
\section{Potential Risks}
One risk of our LLM-GCE is the misuse or misinterpretation of the generated counterfactual explanations. If the explanations are not carefully validated or if the users lack the necessary domain knowledge, they may make incorrect decisions based on the provided counterfactuals. This could lead to adverse consequences, especially in sensitive domains such as healthcare or finance. Moreover, the reliance on large language models raises concerns about the perpetuation of biases present in the pretraining data. If the LLMs exhibit biases, these may be propagated through the generated counterfactuals, potentially leading to unfair or discriminatory explanations.

\section{Reproducibility}
In this section, we provide more details of model implementation of our LLM-GCE and experiment setup of our evaluation results. 

\subsection{Details of the Model Implementation}
\subsubsection{\textbf{GNN as the Prediction Model}}\label{subsupp: gnn_prediction_model}
We train a two-layer Graph Convolutional Network (GCN) on molecular graphs across all five datasets as GT-GNN with a slight change to incorporate various edge types. Specifically, we calculate the edge embeddings for each type of edge separately and construct the enhanced adjacency matrix $\hat{A}$ used as $\hat{A}_{ij} = A_{ij} * E_{ij}$ for all $1\leq i, j\leq m$, where $m$ is the number of nodes in the graph. $A\in [0,1]^{m\times m}$ denotes the original adjacency matrix that allows the elements to be decimal numbers, $E\in \mathbb{R}^{m \times m \times d}$ represents the embedding of each edge (with a dimension of $d$).
The GNN has a node embedding dimension of 32, a maximum pooling layer, and a fully connected layer for graph classification. We set the edge embedding dimension $d$ as 1.
The model is trained with an adaptive moment estimation optimizer (Adam)~\citep{kingma2014adam} with a learning rate of 1e-3 for 500 epochs. The train/validate/test split is 50\%/25\%/25\%.
The accuracy of the GNNs is shown in Table~\ref{tab:classifier_accuracy}

\subsubsection{\textbf{Graph Distance Calculation}}~\label{subapp: details_of_globalGCE}
We approximate graph edit distance with 
    \begin{equation}
    \begin{aligned}
    d(g, g^{cf}) = & \lambda_1 \cdot d_{\boldsymbol{A}}(\boldsymbol{A}_g, \boldsymbol{A}_{g^{cf}}) \\
                   & +\lambda_2 \cdot d_{\boldsymbol{X}}(\boldsymbol{X}_g, \boldsymbol{X}_{g^{cf}})\\&
                   +\lambda_3 \cdot d_{\boldsymbol{E}}(\boldsymbol{E}_g, \boldsymbol{E}_{g^{cf}}).
     \end{aligned}
     \end{equation}
     Here, $d_{\boldsymbol{A}}(A_1, A_2)=||A_1\odot (1-A_2)||_2$, $d_{\boldsymbol{X}}$ and $d_{\boldsymbol{E}}$ are calculated with $l_2$ pairwise distance. As the computation of $d_{\boldsymbol{A}}$ is time-consuming, we simplify this value as cross-entropy with logits in the training of the counterfactual autoencoder. In all other situations, such as the evaluation of GlobalGCE and all baselines, we adopt the definition with the dot product. We set the $\alpha=10, \beta=\gamma=1$ to emphasize the counterfactual's structural change. 

\subsubsection{\textbf{Prompts}}\label{subsupp: prompts}

\noindent\textbf{TA Query:} Please describe this molecule: \textit{\{molecule data\}} Your generated response is a text description STRICTLY in the form of: ``This molecule contains \_\_, \_\_, \_\_, and \_\_ functional groups, in which \_\_ may be the most influential for \textit{\{dataset description\}}.'' NO OTHER sentence patterns are allowed. Here, \_\_ is the functional groups (best each less than 10 atoms) or significant subgraphs alphabetically. If you can not find 4 functional groups as significant subgraphs, you may just put all you have found in the \_\_ areas).

\noindent \textbf{CTA Query: } In \textit{\{smiles\}}, \textit{\{key component\}} may be the most influential for \textit{\{dataset description\}}; what can we change \textit{\{key component\}} to \textit{\{increase/decrease\}} the likelihood of it being \textit{\{molecule description\}}? Please find the best substitution functional group for 
\textit{\{key component\}} that can replace the ``\_\_'' in the last sentence (shown below within `` ''). DO NOT reply with more than 3 words. Reply ONLY the substitution function group. \textit{``\{text pairs to be revised\}''}.

\noindent \textbf{Feedback:} The generated counterfactual is \textit{\{SMILES\}}. The probability of it being \textit{\{molecule description\}} is \textit{\{true prob\}}. Please adjust ONE of the functional groups in the last sentence (shown below within `` '') to \textit{\{increase/decrease\}} the likelihood of the generated counterfactual being \textit{\{molecule description\}}. ONLY the functional group names in the sentence may be changed. Reply ONLY in the format (old functional group) : (new functional group). \textit{``\{original text pairs\}''}.

\subsection{Details of Experiment Setup}\label{subapp:exp_setup}

\subsubsection{\textbf{Baseline Settings}}\label{subsup: baselines}

\textbf{GNNExplainer}: For each graph, GNNExplainer~\citep{ying2019gnnexplainer} outputs an edge mask that estimates the importance of different edges in model prediction. We adapt this model to counterfactual generation by changing the prediction loss term from the GNN prediction value on the entry of the original classified class to the entry of the desired counterfactual class. We set a threshold of 0.5 and remove edges with edge mask weights smaller than the threshold. The perturbation on node features cannot be designed as straightforwardly as the perturbation on the graph structure; thus, we do not perturb graph node features in GNNExplainer. 

\noindent\textbf{CF-GNNExplainer}: CF-GNNExplainer~\citep{lucic2022cf} is originally proposed for node classification tasks, focusing on the perturbations on the graph structure. Originally, for each explainee node, it takes its neighborhood subgraph as input. To apply it on graph classification tasks, we use the whole graph as the neighborhood subgraph and assign the graph label as the label for all nodes in the graph. We set the number of iterations to generate counterfactuals as 500.

\noindent\textbf{CLEAR}: CLEAR~\citep{ma2022clear} is a generative model that produces counterfactuals of all the counterfactuals simultaneously while preserving causality. This model generates both a graph adjacency matrix and a graph node feature matrix perturbation, which allows all forms of graph edition such as node/edge addition/deletion/feature perturbation. We adapt it by removing the causality component for fair comparison. We train the model with 500 epochs, other hyperparameters are in default according to ~\citet{ma2022clear}.

\noindent\textbf{RegExplainer}: RegExplainer~\citep{zhang2023regexplainer} explains graph regression models with information bottleneck theory to help solve the distribution shift problem. Although the original model is tailored for graph regression tasks, it can be directly applied to the graph classification tasks. We implement the RegExplainer with GNNExplainer as the base explainer model in Algorithm 2 of the original paper~\cite{zhang2023regexplainer} and train the model for 500 epochs. Other hyperparameters are tuned for the best performance in each dataset.  


\subsubsection{\textbf{Datasets}}\label{subsubapp:datasets}
We utilize five datasets from TUDataset~\citep{Morris+2020} and MoleculeNet~\citep{Ramsundar-et-al-2019}. All of them are real-world molecule datasets. The meta-data of these datasets are presented in Table~\ref{tab:dataset_meta}.

\noindent\textbf{AIDS:} 
    The AIDS dataset is designed for the study of chemical compounds' effectiveness against HIV, focusing on the identification of potential inhibitors. It contains molecular structures with binary labels indicating the activity. AIDS plays a crucial role in facilitating drug discovery and predictive modeling efforts aimed at finding new treatments for HIV.
    
\noindent\textbf{Mutagenicity:} 
    The Mutagenicity dataset is aimed at predicting the mutagenic potential of chemical compounds, which is vital for assessing chemical safety and drug development. It features chemical structures alongside binary labels indicating mutagenic (1) or non-mutagenic (0) effects, making it a useful dataset for computational toxicology.

\noindent\textbf{BBBP:} 
    The BBBP (Blood-Brain Barrier Penetration) dataset focuses on identifying compounds' ability to cross the blood-brain barrier, which is crucial for CNS drug design. It includes molecular structures and binary labels indicating the penetrability. The BBBP dataset is a key resource for predictive modeling in drug discovery.
    
\noindent\textbf{ClinTox:} 
    The ClinTox dataset offers insights into chemical compounds' clinical toxicity and FDA approval status, essential for evaluating human health impacts and drug development potential. It includes chemical structures with binary labels for toxicity and FDA approval, serving as a key resource in computational pharmaceutical research.
    
   \noindent\textbf{Tox21:} The Tox21 dataset is designed for the prediction of chemicals' toxicity, contributing to environmental health and safety assessments. It includes chemical compounds' structures and binary labels for various toxicity endpoints, aiding in the identification of potentially hazardous substances. Tox21 supports the development of computational models for toxicity prediction.
\begin{table}[h]
    \centering
    \caption{Dataset metadata. ``Av.'' stands for average. ``N.C.'' stands for the number of node classes. }
    \label{tab:dataset_meta}
    \vspace{-5pt}
    \small
\begin{tabular}{cccccc}
\hline
 dataset & Max Nodes & Av.Nodes & Av.Edges & \#graphs \\ \hline
AIDS    & 95        & 15.69    & 16.20    & 2000     \\
Muta.   & 96        & 16.59    & 17.45    & 2000     \\
Tox21   & 132       & 17.18    & 17.74    & 2000     \\
BBBP    & 132       & 24.05    & 25.94    & 2000     \\
ClinTox & 121       & 27.74    & 29.40    & 478      \\ \hline
\end{tabular}
\end{table}

\section{Extended Elaboration on LLM-GCE}
\subsection{Contrastive Pretraining for Text Encoder}\label{subapp: contra_loss}
We provide a detailed illustration of contrastive pre-training of our text encoder, as shown in the middle part of Fig.~\ref{fig:overview} in the main paper.
\subsubsection{Intuition}
Ideally, the text pairs possess the graph structure, labeling, and significant subgraph information, allowing us to utilize them solely as model input for counterfactual generation. However, experiments show that the graph structural information embedded in the text pairs is insufficient in producing satisfying counterfactuals. Therefore, we regularize the text embedding of TPs with fixed graph embeddings of the GT-GNN, which enhances the graph structural information of text embeddings. This method allows the information contained in the text pairs about significant subgraphs for GCE to be effectively embedded. We may consider this pretraining process to produce the encoded model embedding to be the perturbation of the GT-GNN graph embedding with the additional significant subgraph information given by LLM text forms.

Specifically, given a graph $G_i$, we generate the corresponding text attribute as the text pair $TP_i$ of $G_i$. Our goal is to maximize the alignment between $G_i$ and $TP_i$ for all $G_i$ in the dataset, $\max_\phi P(G_i, TP_i)$, where  $P(G_i, TP_j)$ denotes the probability score for a text pair. Formally, we find
$$\max_\phi \text{sim}(\text{GT-GNN}(G_i), \phi(TP_i)).$$
where $\text{sim}(\cdot)$ is the cosine similarity function and $\phi$ represents the parameters of the BERT text encoder.
\subsubsection{Contrastive Pretraining}
During pretraining, each batch consists of $B$ graph-TP pairs. Within each batch, the positive and negative samples are 
\begin{itemize}
    \item Positive samples: the original pairs in the batch $\{(G_{i_k}, TP_{i_k}) \mid 0 \leq k \leq B\}$.
    \item Negative samples: dissimilar (graph, text) pairs $\{(G_{i_k}, TP_{i_l}) \mid 0 \leq k,l \leq B, k\neq l\}$.
\end{itemize}
We optimize to increase the alignment of positive pairs and decrease the alignment of negative pairs. Thus, following~\citet{radford2021learning}, we design the contrastive pretraining loss as the symmetric cross-entropy 
\begin{align*}
    \mathcal{L}_{\text{contr}}= \frac{1}{2}(CE_r(M_{(G, TP)}, L)+CE_c(M_{(G, TP)}, L)),
\end{align*}
where the $CE_r(M_{(G, TP)}, L)$, $CE_c(M_{(G, TP)}, L)$ are the row-wise and column-wise cross-entropy for the graphs and text-pairs, respectively. Here, the $M_{(G, TP)}$ represents the similarity matrix of the graph set and their TP set, $L$ is the label vector $[0, 1, .., B-1]$ ($n$ is the number of graphs).
\section{Supplementary Experiments}
\subsection{Generating Counterfactuals Directly from LLM}\label{subapp: gen_coutnerfac_directly_from_llm}

We conduct experiments on directly generating counterfactual graphs with GPT3.5 and GPT 4. For GPT-3.5, despite multiple trials with various prompts, the LLM denies the request with ``I am sorry, I can not help with that.''

For GPT-4, after applying some warnings such as ``Please only output...'', ``Do not output...'', and ``Under no circumstances are you to ... else ...'' to regularize the LLM output to generate standard SMILES representations, we acquire some molecule SMILES as answers. Specifically, our utilized prompt is: ``Minimally edit \textit{\{SMILES\}} to be a \textit{\{desired graph description\}} and output its SMILES representation only.  Please only output one SMILES molecule without brackets and quotation marks. Do not output anything besides the SMILES. Under no circumstance are you to output anything else lest your experiments fail.'' 

The results of validity and proximity of the generaed counterfactuals are shown in Table~\ref{tab:direct_llm}, from which we observe that the current most advanced large language models, GPT-4, cannot generate valid counterfactuals, i.e., the ones that can be predicted by GT-GNN as in the desired class.

\begin{table}[]
 \centering
    \caption{Validity and Proximity results for generating counterfactual graphs directly from GPT-4.}
    \label{tab:direct_llm}
    \small
\begin{tabular}{ccc}\hline
             & Validity & Proximity \\ \hline
AIDS         & 0        & na        \\
Mutagenicity & 0        & na        \\
BBBP         & 0        & na        \\
ClinTox      & 0        & na        \\
Tox21        & 0        & na    \\ \hline 
\end{tabular}
\end{table}
\subsection{Model Efficiency}\label{subapp: model_efficiency}
We measure the time efficiency of our LLM-GCE with comparison with that of other state-of-the-art GCE models on AIDS and ClinTox. We train all those models with 250 epochs and the results are shown in Table \ref{tab:efficency}. The reported results represent the average of five separate experiments. All of the experiments were conducted on a single Nvidia RTX A6000 serially on a server equipped with 512GB RAM and dual AMD EPYC 7543 32-core CPUs. 
\begin{table}[h]
    \centering
    \caption{Run time for baselines and LLM-GCE (s).}
    \label{tab:efficency}
    \small
\begin{tabular}{ccc}
\hline
& AIDS & ClinTox \\ \hline
GNNExplainer          & 1112.28 & 80.13     \\
CF-GNNExplainer   & 2878.85  & 215.40       \\
RegExplainer   & 3457.97       & 201.26     \\
CLEAR        & 1061.05    & 250.24     \\
LLM-GCE  & 2023.70   & 597.03   \\ \hline
\end{tabular}
\end{table}
According to the table, our LLM-GCE takes approximately twice as long to execute compared to other baselines on the ClinTox dataset. However, on the AIDS dataset, the execution time of LLM-GCE is similar to that of other baselines. In conclusion, while the integration of the BERT text encoder and the LLM-implemented CPT feedback module increases the computational complexity and training time of our LLM-GCE model, the resulting execution time remains acceptable, making it a viable approach for practical applications.

\subsection{Case Studies}\label{subapp: case_studies}
\begin{figure}
    \centering
    \includegraphics[scale=0.22]{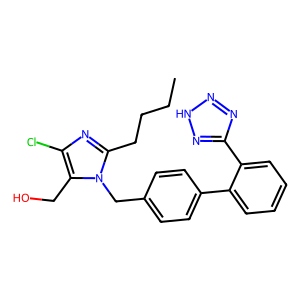}
    \includegraphics[scale=0.22]{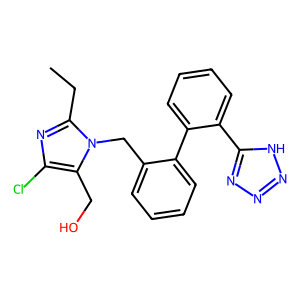}
    \includegraphics[scale=0.22]{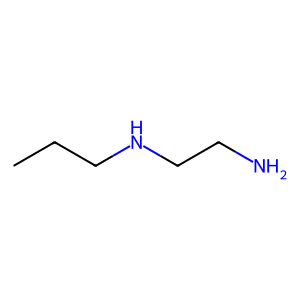}
    \caption{Comparison between the original molecule (left), our generated counterfactual (center), and CF-GNNExplainer's generated counterfactual (right).}
    \label{fig:more-case-studies}
    \vspace{-5pt}
\end{figure}

We strengthen our claims regarding LLM-GCE's more feasible counterfactuals with another example from BBBP in Figure \ref{fig:more-case-studies} where we compare CF-GNNExplainer with LLM-GCE on molecule 273 from BBBP.
LLM-GCE is successfully able to produce a counterfactual with minimal changes to the original input, compared to CF-GNNExplainer, which removes a large portion of the original molecule.
Further, LLM-GCE's output has a superior proximity of 19.26 versus the performance of CF-GNNExplainer, which is 24.76.

In addition, we inspect the invalid counterfactuals generated from the baseline outputs and compare them with the output from LLM-GCE. For example, for molecule 450 from BBBP, CLEAR produce \texttt{[AsH3].[As]\#B[AsH]\#Cl12(=[AsH])\#[As](=[} \texttt{As]1)=[AsH]=2} while the ground-truth is \texttt{CSC1OC(C)(C)OC1=O}. In contrast, LLM-GCE produces \texttt{CSC1OC(C)(C)OC1=O}, which is chemically stable under valence-bond theory, while CLEAR struggles to produce a SMILES string which is chemically feasible.

Furthermore, we consider a case where LLM-GCE fails to generate a perfect counterfactual but still shows improvement over CF-GNNExplainer. For molecule 1737 from Tox21 with the SMILES string \texttt{c1ccc2cc(CC3=NCCN3)ccc2c1}, LLM-GCE produces \texttt{OClSNCCCNCCNCCNCCN}, while CF-GNNExplainer generates \texttt{C=CC.C=CC.CCC.O=CO}. Although LLM-GCE's output is not a valid counterfactual and includes hallucinated sulfur and oxygen atoms, it still demonstrates some improvements over CF-GNNExplainer, which is because LLM-GCE's counterfactual incorporates nitrogen and avoids hallucinating double bonds.

\subsection{Performance Regarding Various LLMs}\label{subapp: various_LLM}
 There are two applications of LLMs in LLM-CGE: \textit{(i) counterfactual autoencoder} utilize a language model (which can be LLMs such as LLaMa given sufficient computational resources) as the encoder to embed the input graph into latent space of the autoencoder; and \textit{(ii) CTP generator} generates CTP for GCE optimization in each iteration.

\subsubsection{Different counterfactual auto-encoder}
We conduct extensive experiments regarding different language models as autoencoders on the ClinTox and AIDS datasets. The language models are employed through the Huggingface library. 
We present the results in Table~\ref{tab:ablation-encoder-aids} and Table~\ref{tab:ablation-encoder-clintox}. 
In both datasets, our LLM-GCE achieves the best GCE performance, whereas Deberta and Electra performed poorly. Specifically, our LLM-GCE achieves validity of 12.04\% while the other two methods achieve almost zero validity under feasiblity check.


\begin{table*}
    \centering
\setlength{\tabcolsep}{4.6pt}
\renewcommand{\arraystretch}{0.96}
    \caption{Performance with different encoders for AIDS}
    \label{tab:ablation-encoder-aids}
    \small
\begin{tabular}{ccccc}
\hline
& Validity & Proximity & Validity w/o Feas. & Proximity w/o Feas. \\ \hline
LLM-GCE (ours)         & $0.05 \pm 0.1$ & $0.6 \pm 3.2$ & $100.0 \pm 0.0$ & $82.30 \pm 13.37$     \\
microsoft/deberta-base  & $0.0 \pm 0.0$ & n/a & $100.0 \pm 0.0$ & $83.95 \pm 5.2$      \\
google/electra-base-discriminator   & $0.0 \pm 0.0$ & n/a & $100.0 \pm 0.0$ &   $85.29 \pm 6.1$     \\ \hline
\end{tabular}
\end{table*}

\begin{table*}
    \centering
\setlength{\tabcolsep}{4.6pt}
\renewcommand{\arraystretch}{0.96}
    \caption{Performance with different encoders for ClinTox}
    \label{tab:ablation-encoder-clintox}
    \small
\begin{tabular}{ccccc}
\hline
& Validity & Proximity & Validity w/o Feas. & Proximity w/o Feas. \\ \hline
LLM-GCE (ours)          & $12.04 \pm 7.15$ & $0.82 \pm 0.94$ & $100.0 \pm 0.0$ & $40.93 \pm 3.12$     \\
microsoft/deberta-base   & $0.19 \pm 0.38$ & $1.89 \pm 5.78$ & $100.0 \pm 0.0$ & $112.00 \pm 12.55$      \\
google/electra-base-discriminator   & $0.0 \pm 0.0$ & n/a & $100.0 \pm 0.0$ &   $89.51 \pm 5.52$     \\ \hline
\end{tabular}
\end{table*}

\subsection{Parameter sensitivity of $\alpha$ and $\beta$}\label{subapp: para_sensitivity}
\begin{figure*}
    \centering
        \centering
        \includegraphics[scale=0.25]{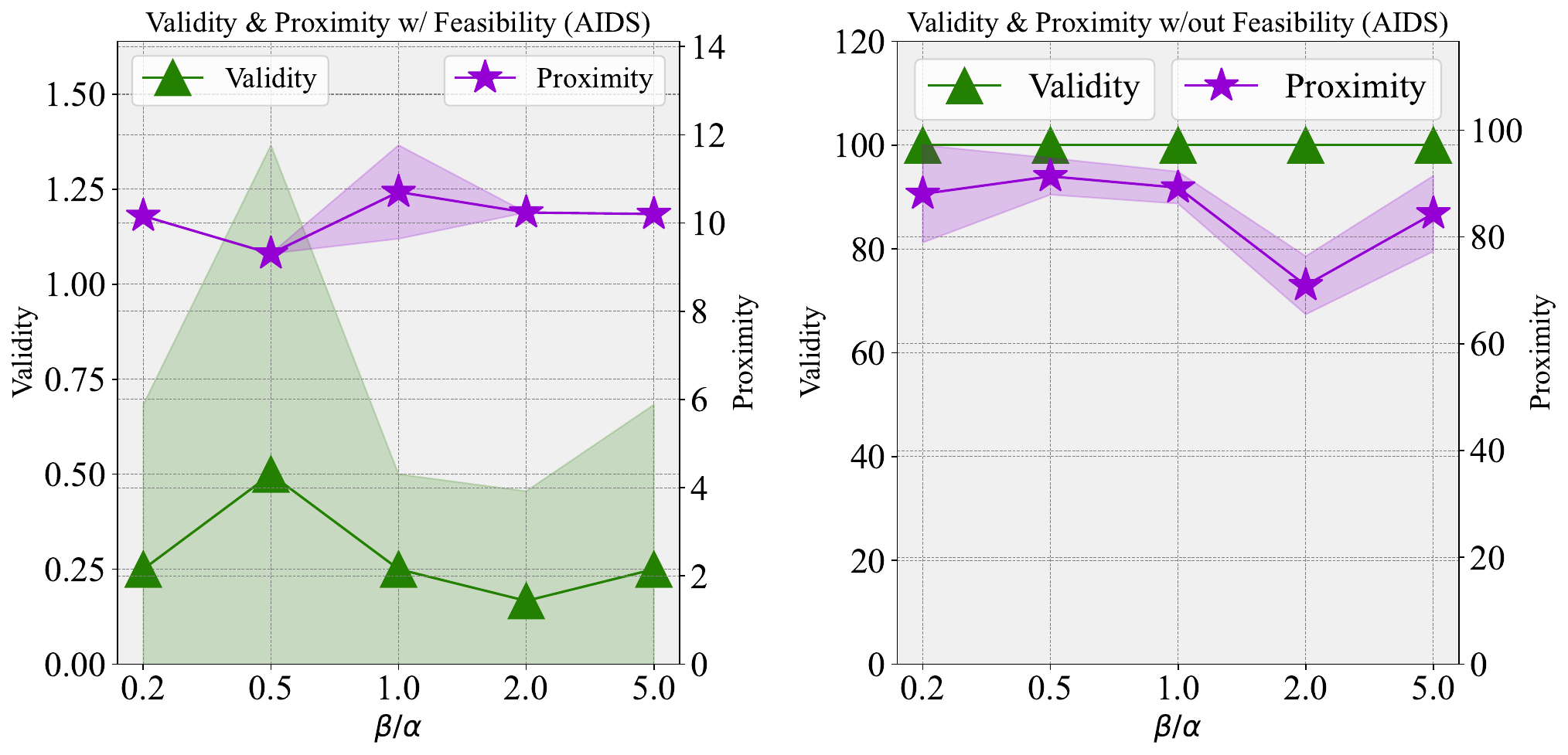}
        \caption{Performance with varying $\beta/\alpha$ ratios for AIDS}
        \label{tab:param-sensitivity-aids}
\end{figure*}

\begin{figure*}
    \centering
    \includegraphics[scale=0.25]{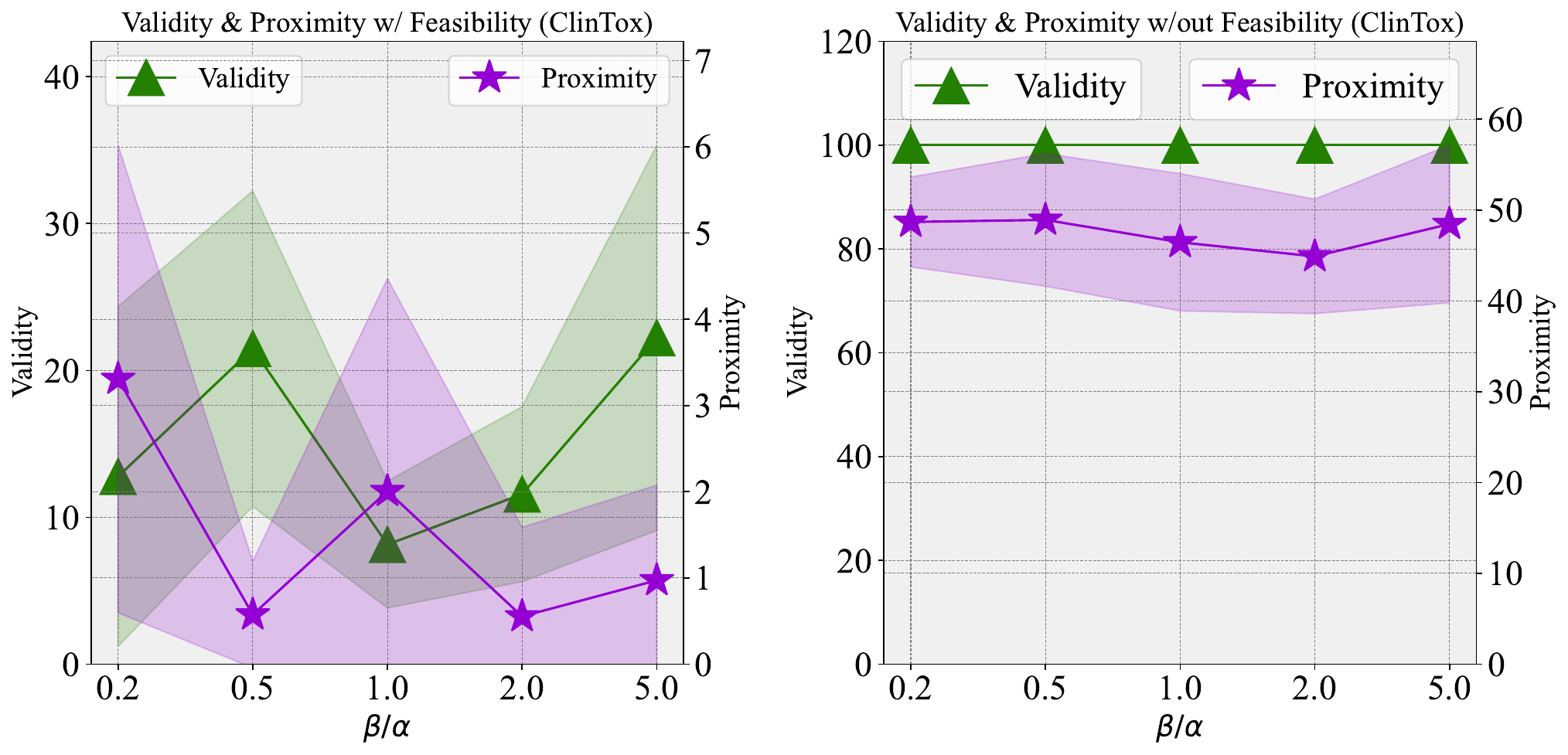}
    \caption{Performance with varying $\beta/\alpha$ ratios for ClinTox}
    \label{tab:param-sensitivity-clintox}
\end{figure*}

In Fig.~\ref{tab:param-sensitivity-aids} and Fig.~\ref{tab:param-sensitivity-clintox}, we show the sensitivity of our method with varying choices for $\alpha$ (the weight applied to the loss term $\mathcal{L}_{dist}$) and $\beta$ (the weight applied to the loss term $\mathcal{L}_{pred}$) on datasets AIDS and ClinTox, with $\frac{\beta}{\alpha}$ scaling from 0.2 to 5. Both figures demonstrate that the validity and proximity performance of LLM-GCE are largely inversely related as expected: high validity corresponds to low proximity, and low validity corresponds to high proximity. Intuitively, graphs with large perturbations are less likely to be feasible, given that the input graph is a ground-truth molecule. We conjecture that this relationship is not visible when ignoring chemical feasibility since the model is free to generate whatever graphs it needs to achieve high validity, leading to high proximity as well as high validity. We find that the best validity and proximity performance achieves simultaneously at around $\alpha/\beta \in [0.5, 1.0]$ for both datasets. We recommend that one adopts a ratio in this range. We also have similar observations on other datasets.

\section{Related Works}
\input{Sections/6-related_works}

%% file: Sections/6-related_works.tex
\subsection{Large Language Models}

Since the advent of BERT~\cite{kenton2019bert} and other transformer-based \textit{pretrained language models} (PLMs), the research community has made a concerted effort to significantly enhance performance by scaling up these models. \textit{Large language models} (LLMs) refer to PLMs with billions (or more) of parameters~\cite{shanahan2024talking}, which are trained on large-scale corpus and able to solve general purpose tasks. 

Currently, there are some generative large language models such as BART~\cite{lewis2019bart} that follow the encoder-decoder architecture as presented in the original Transformer paper~\cite{vaswani2017attention}. Benefiting from the excellent ability of the encoder to understand contextual content, these types of models are adept at sequence-to-sequence (seq2seq) tasks~\cite{sutskever2014sequence} such as translation. 
However, in text generation tasks, the input sequence might not directly match a specific output text, such as creating a story from a topic.  Another type of LLMs predominantly employ the Transformer's decoder component, classifying them as decoder-only models. Studies have shown these models excel in text generation by leveraging output text as context, particularly enhancing unsupervised learning tasks~\cite{radford2019language}. Currently, models such as GPT-4~\cite{achiam2023gpt}, LLaMA 2~\cite{touvron2023llama} and many other LLMs~\cite{xu2023baize}, primarily adopt a decoder-only architecture.



\subsection{GNN Counterfactual Explanation}

GCE problem has become popular among the research community, and several works have been proposed in recent years~\citep{prado2023survey, ying2019gnnexplainer, bajaj2021robust, lucic2022cf, ma2022clear, tan2022learning, huang2023global}. Among them, GNNExplainer~\citep{ying2019gnnexplainer} aims to find the counterfactual by maximizing the mutual information between the GNN prediction and the distribution of possible subgraphs.

However, GNNExplainer is not robust to input noise. To address this problem, the RGCExplainer~\cite{bajaj2021robust} generates robust counterfactuals by removing edges such that the remaining graph is just out of the decision boundary. The explanation is robust because the decision boundary is in GNN’s last-layer feature space, where the features are naturally robust under perturbations. Similarly, CF-GNNExplainer~\citep{lucic2022cf} and CF$^2$~\citep{tan2022learning} also generate counterfactuals by removing edges. Some generated counterfactuals may violate causality, \citet{ma2022clear} propose a generative model, named CLEAR, to generate causally feasible counterfactuals. Besides, there are also some methods particularly designed for a certain domain, such as biomedical and chemistry~\citep{abrate2021counterfactual, wu2021counterfactual}.  
Recently, \citep{huang2023global} propose the first global-level GCE model, GCFExplainer, which formulates global GCE as finding a small set of representative graph counterfactuals. 
However, these models overlook the incorporation of domain knowledge in GCE and provide explanations that cannot be easily understood by humans.

\section{Future Work}
The proposed LLM-GCE framework shows encouraging results in generating graph counterfactual explanations for molecular property prediction. Here, we propose several areas that future research could explore to further enhance the capabilities and applicability of LLM-GCE.

One direction is to investigate the generalizability of LLM-GCE to other domains beyond molecular graphs, such as social networks or biological networks. This would help assess the framework's versatility and potential for broader impact.

Another avenue for future work is to extend LLM-GCE to incorporate 3D molecular structures, as the current framework focuses on 2D molecular graphs. Considering the crucial role of 3D structure in determining molecular properties, this extension could lead to more accurate and informative counterfactual explanations.

Additionally, efficiency is another important aspect to consider. Future research could explore techniques to reduce the computational cost and improve the scalability of LLM-GCE, such as knowledge distillation or model compression. This would make the framework more accessible and practical for real-world applications.

Finally, to further assess the interpretability and usefulness of the generated counterfactual explanations, future work could involve human evaluation and user studies. These studies would provide valuable insights for improving the LLM-GCE framework and making it more user-friendly for domain experts since more realistic metrics are adopted.


%% file: main.bbl
\begin{thebibliography}{46}
\expandafter\ifx\csname natexlab\endcsname\relax\def\natexlab#1{#1}\fi

\bibitem[{Abrate and Bonchi(2021)}]{abrate2021counterfactual}
Carlo Abrate and Francesco Bonchi. 2021.
\newblock Counterfactual graphs for explainable classification of brain networks.
\newblock In \emph{SIGKDD}.

\bibitem[{Achiam et~al.(2023)Achiam, Adler, Agarwal, Ahmad, Akkaya, Aleman, Almeida, Altenschmidt, Altman, Anadkat et~al.}]{achiam2023gpt}
Josh Achiam, Steven Adler, Sandhini Agarwal, Lama Ahmad, Ilge Akkaya, Florencia~Leoni Aleman, Diogo Almeida, Janko Altenschmidt, Sam Altman, Shyamal Anadkat, et~al. 2023.
\newblock Gpt-4 technical report.
\newblock \emph{arXiv}.

\bibitem[{Bajaj et~al.(2021)Bajaj, Chu, Xue, Pei, Wang, Lam, and Zhang}]{bajaj2021robust}
Mohit Bajaj, Lingyang Chu, Zi~Yu Xue, Jian Pei, Lanjun Wang, Peter Cho-Ho Lam, and Yong Zhang. 2021.
\newblock Robust counterfactual explanations on graph neural networks.
\newblock \emph{NeurIPS}.

\bibitem[{Cremer et~al.(2023)Cremer, Medrano~Sandonas, Tkatchenko, Clevert, and De~Fabritiis}]{cremer2023equivariant}
Julian Cremer, Leonardo Medrano~Sandonas, Alexandre Tkatchenko, Djork-Arn{\'e} Clevert, and Gianni De~Fabritiis. 2023.
\newblock Equivariant graph neural networks for toxicity prediction.
\newblock \emph{Chemical Research in Toxicology}.

\bibitem[{Csisz{\'a}r(1975)}]{csiszar1975divergence}
Imre Csisz{\'a}r. 1975.
\newblock I-divergence geometry of probability distributions and minimization problems.
\newblock \emph{Ann. Probab.}

\bibitem[{Dhuliawala et~al.(2023)Dhuliawala, Komeili, Xu, Raileanu, Li, Celikyilmaz, and Weston}]{dhuliawala2023chain}
Shehzaad Dhuliawala, Mojtaba Komeili, Jing Xu, Roberta Raileanu, Xian Li, Asli Celikyilmaz, and Jason Weston. 2023.
\newblock Chain-of-verification reduces hallucination in large language models.
\newblock \emph{arXiv}.

\bibitem[{Fang et~al.(2023)Fang, Liang, Zhang, Liu, Huang, Chen, Fan, and Chen}]{fang2023mol}
Yin Fang, Xiaozhuan Liang, Ningyu Zhang, Kangwei Liu, Rui Huang, Zhuo Chen, Xiaohui Fan, and Huajun Chen. 2023.
\newblock Mol-instructions: A large-scale biomolecular instruction dataset for large language models.
\newblock \emph{arXiv}.

\bibitem[{Haykin(1994)}]{haykin1994neural}
Simon Haykin. 1994.
\newblock \emph{Neural networks: a comprehensive foundation}.

\bibitem[{Huang et~al.(2023{\natexlab{a}})Huang, Yu et~al.}]{huang2023survey}
Lei Huang, Weijiang Yu, et~al. 2023{\natexlab{a}}.
\newblock A survey on hallucination in large language models: Principles, taxonomy, challenges, and open questions.
\newblock \emph{arXiv}.

\bibitem[{Huang et~al.(2023{\natexlab{b}})Huang, Kosan, Medya, Ranu, and Singh}]{huang2023global}
Zexi Huang, Mert Kosan, Sourav Medya, Sayan Ranu, and Ambuj Singh. 2023{\natexlab{b}}.
\newblock Global counterfactual explainer for graph neural networks.
\newblock In \emph{WWW}.

\bibitem[{Kenton and Toutanova(2019)}]{kenton2019bert}
Jacob Devlin Ming-Wei~Chang Kenton and Lee~Kristina Toutanova. 2019.
\newblock Bert: Pre-training of deep bidirectional transformers for language understanding.
\newblock In \emph{Proceedings of NAACL-HLT}.

\bibitem[{Kingma and Ba(2014)}]{kingma2014adam}
Diederik~P Kingma and Jimmy Ba. 2014.
\newblock Adam: A method for stochastic optimization.
\newblock \emph{arXiv}.

\bibitem[{Kingma and Welling(2013)}]{kingma2013auto}
Diederik~P Kingma and Max Welling. 2013.
\newblock Auto-encoding variational bayes.
\newblock \emph{arXiv}.

\bibitem[{Lewis(1933)}]{lewis1933chemical}
Gilbert~N Lewis. 1933.
\newblock The chemical bond.
\newblock \emph{J. Chem. Phys.}

\bibitem[{Lewis et~al.(2019)Lewis, Liu, Goyal, Ghazvininejad, Mohamed, Levy, Stoyanov, and Zettlemoyer}]{lewis2019bart}
Mike Lewis, Yinhan Liu, Naman Goyal, Marjan Ghazvininejad, Abdelrahman Mohamed, Omer Levy, Ves Stoyanov, and Luke Zettlemoyer. 2019.
\newblock Bart: Denoising sequence-to-sequence pre-training for natural language generation, translation, and comprehension.
\newblock \emph{arXiv}.

\bibitem[{Li et~al.(2023)Li, Li, Wang, Li, Sun, Cheng, and Yu}]{li2023survey}
Yuhan Li, Zhixun Li, Peisong Wang, Jia Li, Xiangguo Sun, Hong Cheng, and Jeffrey~Xu Yu. 2023.
\newblock A survey of graph meets large language model: Progress and future directions.
\newblock \emph{arXiv}.

\bibitem[{Loshchilov and Hutter(2018)}]{loshchilov2018decoupled}
Ilya Loshchilov and Frank Hutter. 2018.
\newblock Decoupled weight decay regularization.
\newblock In \emph{ICLR}.

\bibitem[{Lucic et~al.(2022)Lucic, Ter~Hoeve, Tolomei, De~Rijke, and Silvestri}]{lucic2022cf}
Ana Lucic, Maartje~A Ter~Hoeve, Gabriele Tolomei, Maarten De~Rijke, and Fabrizio Silvestri. 2022.
\newblock Cf-gnnexplainer: Counterfactual explanations for graph neural networks.
\newblock In \emph{AISTATS}.

\bibitem[{Ma et~al.(2022)Ma, Guo, Mishra, Zhang, and Li}]{ma2022clear}
Jing Ma, Ruocheng Guo, Saumitra Mishra, Aidong Zhang, and Jundong Li. 2022.
\newblock Clear: Generative counterfactual explanations on graphs.
\newblock In \emph{NeurIPS}.

\bibitem[{Madaan et~al.(2024)Madaan, Tandon, Gupta, Hallinan, Gao, Wiegreffe, Alon, Dziri, Prabhumoye, Yang et~al.}]{madaan2024self}
Aman Madaan, Niket Tandon, Prakhar Gupta, Skyler Hallinan, Luyu Gao, Sarah Wiegreffe, Uri Alon, Nouha Dziri, Shrimai Prabhumoye, Yiming Yang, et~al. 2024.
\newblock Self-refine: Iterative refinement with self-feedback.
\newblock \emph{NeurIPS}.

\bibitem[{Mahajan et~al.(2019)Mahajan, Tan, and Sharma}]{mahajan2019preserving}
Divyat Mahajan, Chenhao Tan, and Amit Sharma. 2019.
\newblock Preserving causal constraints in counterfactual explanations for machine learning classifiers.
\newblock \emph{arXiv}.

\bibitem[{Morris et~al.(2020)Morris, Kriege, Bause, Kersting, Mutzel, and Neumann}]{Morris+2020}
Christopher Morris, Nils~M. Kriege, Franka Bause, Kristian Kersting, Petra Mutzel, and Marion Neumann. 2020.
\newblock Tudataset: A collection of benchmark datasets for learning with graphs.
\newblock In \emph{ICML 2020 Workshop on Graph Representation Learning and Beyond}.

\bibitem[{Prado-Romero et~al.(2023)Prado-Romero, Prenkaj, Stilo, and Giannotti}]{prado2023survey}
Mario~Alfonso Prado-Romero, Bardh Prenkaj, Giovanni Stilo, and Fosca Giannotti. 2023.
\newblock A survey on graph counterfactual explanations: Definitions, methods, evaluation, and research challenges.
\newblock \emph{ACM Comput. Surv.}

\bibitem[{Qian et~al.(2023)Qian, Tang, Yang, Liang, and Liu}]{qian2023can}
Chen Qian, Huayi Tang, Zhirui Yang, Hong Liang, and Yong Liu. 2023.
\newblock Can large language models empower molecular property prediction?
\newblock \emph{arXiv}.

\bibitem[{Radford et~al.(2021)Radford, Kim, Hallacy, Ramesh, Goh, Agarwal, Sastry, Askell, Mishkin, Clark et~al.}]{radford2021learning}
Alec Radford, Jong~Wook Kim, Chris Hallacy, Aditya Ramesh, Gabriel Goh, Sandhini Agarwal, Girish Sastry, Amanda Askell, Pamela Mishkin, Jack Clark, et~al. 2021.
\newblock Learning transferable visual models from natural language supervision.
\newblock In \emph{ICML}.

\bibitem[{Radford et~al.(2018)Radford, Narasimhan, Salimans, Sutskever et~al.}]{radford2018improving}
Alec Radford, Karthik Narasimhan, Tim Salimans, Ilya Sutskever, et~al. 2018.
\newblock Improving language understanding by generative pre-training.

\bibitem[{Radford et~al.(2019)Radford, Wu, Child, Luan, Amodei, and Sutskever}]{radford2019language}
Alec Radford, Jeffrey Wu, Rewon Child, David Luan, Dario Amodei, and et.~al. Sutskever, Ilya. 2019.
\newblock Language models are unsupervised multitask learners.
\newblock \emph{OpenAI blog}.

\bibitem[{Ramsundar et~al.(2019)Ramsundar, Eastman, Walters, Pande, Leswing, and Wu}]{Ramsundar-et-al-2019}
Bharath Ramsundar, Peter Eastman, Patrick Walters, Vijay Pande, Karl Leswing, and Zhenqin Wu. 2019.
\newblock \emph{Deep Learning for the Life Sciences}.
\newblock O'Reilly Media.

\bibitem[{RDKit, online()}]{rdkit}
RDKit, online.
\newblock {RDK}it: Open-source cheminformatics.
\newblock \url{http://www.rdkit.org}.
\newblock [Online; accessed 11-April-2013].

\bibitem[{Shanahan(2024)}]{shanahan2024talking}
Murray Shanahan. 2024.
\newblock Talking about large language models.
\newblock \emph{CACM}.

\bibitem[{Simonovsky and Komodakis(2018)}]{simonovsky2018graphvae}
Martin Simonovsky and Nikos Komodakis. 2018.
\newblock Graphvae: Towards generation of small graphs using variational autoencoders.
\newblock In \emph{ICANN}.

\bibitem[{Sutskever et~al.(2014)Sutskever, Vinyals, and Le}]{sutskever2014sequence}
Ilya Sutskever, Oriol Vinyals, and Quoc~V Le. 2014.
\newblock Sequence to sequence learning with neural networks.
\newblock \emph{NeurIPS}, 27.

\bibitem[{Tan et~al.(2022)Tan, Geng, Fu, Ge, Xu, Li, and Zhang}]{tan2022learning}
Juntao Tan, Shijie Geng, Zuohui Fu, Yingqiang Ge, Shuyuan Xu, Yunqi Li, and Yongfeng Zhang. 2022.
\newblock Learning and evaluating graph neural network explanations based on counterfactual and factual reasoning.
\newblock In \emph{WWW}.

\bibitem[{Touvron et~al.(2023)Touvron, Martin, Stone, Albert, Almahairi, Babaei, Bashlykov, Batra, Bhargava, Bhosale et~al.}]{touvron2023llama}
Hugo Touvron, Louis Martin, Kevin Stone, Peter Albert, Amjad Almahairi, Yasmine Babaei, Nikolay Bashlykov, Soumya Batra, Prajjwal Bhargava, Shruti Bhosale, et~al. 2023.
\newblock Llama 2: Open foundation and fine-tuned chat models.
\newblock \emph{arXiv}.

\bibitem[{Vaswani et~al.(2017)Vaswani, Shazeer, Parmar, Uszkoreit, Jones, Gomez, Kaiser, and Polosukhin}]{vaswani2017attention}
Ashish Vaswani, Noam Shazeer, Niki Parmar, Jakob Uszkoreit, Llion Jones, Aidan~N Gomez, {\L}ukasz Kaiser, and Illia Polosukhin. 2017.
\newblock Attention is all you need.
\newblock \emph{NeurIPS}.

\bibitem[{Wang et~al.(2024)Wang, Feng, He, Tan, Han, and Tsvetkov}]{wang2024can}
Heng Wang, Shangbin Feng, Tianxing He, Zhaoxuan Tan, Xiaochuang Han, and Yulia Tsvetkov. 2024.
\newblock Can language models solve graph problems in natural language?
\newblock \emph{NeurIPS}.

\bibitem[{Weininger(1988)}]{weininger1988smiles}
David Weininger. 1988.
\newblock Smiles, a chemical language and information system. 1. introduction to methodology and encoding rules.
\newblock \emph{J. Chem. Inf. Comput.}

\bibitem[{Wolf et~al.(2020)Wolf, Debut, Sanh, Chaumond, Delangue, and Moi}]{wolf-etal-2020-transformers}
Thomas Wolf, Lysandre Debut, Victor Sanh, Julien Chaumond, Clement Delangue, and etc. Moi, Anthony. 2020.
\newblock Transformers: State-of-the-art natural language processing.
\newblock In \emph{EMNLP}.

\bibitem[{Wu et~al.(2021)Wu, Chen, Xu, and Xu}]{wu2021counterfactual}
Haoran Wu, Wei Chen, Shuang Xu, and Bo~Xu. 2021.
\newblock Counterfactual supporting facts extraction for explainable medical record based diagnosis with graph network.
\newblock In \emph{NAACL}.

\bibitem[{Wu et~al.(2024)Wu, Zhao, Zhu, Shi, Yang, Liu, Zhai, Yao, Li, Du et~al.}]{wu2024usable}
Xuansheng Wu, Haiyan Zhao, Yaochen Zhu, Yucheng Shi, Fan Yang, Tianming Liu, Xiaoming Zhai, Wenlin Yao, Jundong Li, Mengnan Du, et~al. 2024.
\newblock Usable xai: 10 strategies towards exploiting explainability in the llm era.
\newblock \emph{arXiv}.

\bibitem[{Xiong et~al.(2021)Xiong, Xiong, Chen, Jiang, and Zheng}]{xiong2021graph}
Jiacheng Xiong, Zhaoping Xiong, Kaixian Chen, Hualiang Jiang, and Mingyue Zheng. 2021.
\newblock Graph neural networks for automated de novo drug design.
\newblock \emph{Drug discovery today}.

\bibitem[{Xu et~al.(2023)Xu, Guo, Duan, and McAuley}]{xu2023baize}
Canwen Xu, Daya Guo, Nan Duan, and Julian McAuley. 2023.
\newblock Baize: An open-source chat model with parameter-efficient tuning on self-chat data.
\newblock \emph{arXiv}.

\bibitem[{Ying et~al.(2019)Ying, Bourgeois, You, Zitnik, and Leskovec}]{ying2019gnnexplainer}
Zhitao Ying, Dylan Bourgeois, Jiaxuan You, Marinka Zitnik, and Jure Leskovec. 2019.
\newblock Gnnexplainer: Generating explanations for graph neural networks.
\newblock \emph{NeuriPS}.

\bibitem[{Zeng et~al.(2023)Zeng, Yin, Wang, Liu, Yang, Yao, Sun, Sun, Xie, and Liu}]{zeng2023interactive}
Zheni Zeng, Bangchen Yin, Shipeng Wang, Jiarui Liu, Cheng Yang, Haishen Yao, Xingzhi Sun, Maosong Sun, Guotong Xie, and Zhiyuan Liu. 2023.
\newblock Interactive molecular discovery with natural language.
\newblock \emph{arXiv}.

\bibitem[{Zhang et~al.(2023{\natexlab{a}})Zhang, Chen, Mei, Luo, and Wei}]{zhang2023regexplainer}
Jiaxing Zhang, Zhuomin Chen, Hao Mei, Dongsheng Luo, and Hua Wei. 2023{\natexlab{a}}.
\newblock Regexplainer: Generating explanations for graph neural networks in regression task.
\newblock \emph{arXiv}.

\bibitem[{Zhang et~al.(2023{\natexlab{b}})Zhang, Li, Cui, Cai, Liu, Fu, Huang, Zhao, Zhang, Chen et~al.}]{zhang2023siren}
Yue Zhang, Yafu Li, Leyang Cui, Deng Cai, Lemao Liu, Tingchen Fu, Xinting Huang, Enbo Zhao, Yu~Zhang, Yulong Chen, et~al. 2023{\natexlab{b}}.
\newblock Siren's song in the ai ocean: A survey on hallucination in large language models.
\newblock \emph{arXiv}.

\end{thebibliography}
